\documentclass{article}

\usepackage{arxiv}

\usepackage[utf8]{inputenc} 
\usepackage[T1]{fontenc}    
\usepackage{hyperref}       
\usepackage{url}            
\usepackage{booktabs}       
\usepackage{amsfonts}       
\usepackage{nicefrac}       
\usepackage{microtype}      
\usepackage{lipsum}

\usepackage{multirow}
\usepackage{algorithm}
\usepackage{algorithmic}
\usepackage{setspace}
\usepackage{bm}
\usepackage{graphicx}
\usepackage{enumerate}
\usepackage{balance}
\usepackage{color}
\usepackage{epstopdf}
\usepackage{subfigure}
\usepackage{url}
\usepackage{wrapfig,epsf}
\usepackage{dsfont}

\title{Autoregressive-Model-Based Methods for Online Time Series Prediction with Missing Values: an Experimental Evaluation}

\author{
  Xi Chen \\
  School of Computer Science and Technology\\
  Harbin Institute of Technology\\
  Harbin, P.R.China \\
  \texttt{18s003022@stu.hit.edu.cn} \\
   \And
 Hongzhi Wang \\
  School of Computer Science and Technology\\
  Harbin Institute of Technology\\
  Harbin, P.R.China \\
  \texttt{wangzh@hit.edu.cn} \\
   \AND
   Yanjie Wei \\
   School of Computer Science and Technology \\
   Harbin Institute of Technology \\
   Harbin, P.R.China \\
   \texttt{weiyanjie10@gmail.com} \\
   \And
   Jianzhong Li \\
   School of Computer Science and Technology \\
   Harbin Institute of Technology \\
   Harbin, P.R.China \\
   \texttt{lijzh@hit.edu.cn} \\
   \And
   Hong Gao \\
   School of Computer Science and Technology \\
   Harbin Institute of Technology \\
    Harbin, P.R.China \\
   \texttt{honggao@hit.edu.cn} \\
}

\begin{document}
\maketitle

\begin{abstract}
Time series prediction with missing values is an important problem of time series analysis since complete data is usually hard to obtain in many real-world applications. To model the generation of time series, autoregressive (AR) model is a basic and widely used one, which assumes that each observation in the time series is a noisy linear combination of some previous observations along with a constant shift. To tackle the problem of prediction with missing values, a number of methods were proposed based on various data models. For real application scenarios, how do these methods perform over different types of time series with different levels of data missing remains to be investigated. In this paper, we focus on online methods for AR-model-based time series prediction with missing values. We adapted five mainstream methods to fit in such a scenario. We make detailed discussion on each of them by introducing their core ideas about how to estimate the AR coefficients and their different strategies to deal with missing values. We also present algorithmic implementations for better understanding. In order to comprehensively evaluate these methods and do the comparison, we conduct experiments with various configurations of relative parameters over both synthetic and real data. From the experimental results, we derived several noteworthy conclusions and shows that imputation is a simple but reliable strategy to handle missing values in online prediction tasks.
\end{abstract}


\section{INTRODUCTION}
Typically, a time series is a sequence of  data collected at successive intervals of time. For example, atmospheric temperature and stock price are typical examples of time series. Since time series pervasively exist in human and natural activities, it is necessary to build models for them. With the help of these models, meaningful statistics and other characteristics of the data can be extracted through time series analysis.

In such a research area, time series prediction has long been an important problem. Its primary goal is to make good predictions about future data in the time series based on previous data.
In many applications, the prediction results are helpful for decision making. With the good insight of data trend, a system cannot only make better adjustment accordingly but also avoid potential catastrophes. Therefore, methods for better prediction results are in demand.

A great deal of models and algorithms have been proposed in order to achieve more accurate prediction or accelerate computation. However, many of them encountered difficulty on real-world data-centric applications. One important reason is that there are missing values in real-world time series. Such data loss is very common because of device malfunction or channel disturbance. Faced with missing values, traditional methods need to wait for the repair of data before continuing the prediction process. However, this can be improper sometimes.

For instance, a widely used technique for RTT (Round Trip Time) estimation \cite{karn1987improving} in computer networks is called Exponentially Weighted Moving Average (EWMA). In order to compute the estimation, the current observation of RTT is required. However, if the data is lost, the traditional methods have to remeasure it in extra time, during which the network state may change to make the repaired observation not reliable. This can make EWMA unable to track the latest state of the network congestion, which may result in hypofunction of congestion control in reliable data transmission. Therefore, the limitation of traditional methods in practical use is conspicuous, and developing missing-value-tolerant methods is a current trend.

To handle the problem of missing values, there are basically two ways of thinking. One is to impute the missing values before applying the traditional prediction methods. Such an approach can be easy to implement, since we only need to use the previous prediction results to fill in the missing positions. In this case, classic methods like Yule-Walker Equation (YW) \cite{eshel2003yule}, Online Gradient Descent (OGD) \cite{ying2008online} and Kalman Filter (KF) \cite{kalman1960new} can be easily adapted to become missing value tolerant. YW focuses on the autocorrelations in time series and compute the weight vector by solving the Yule-Walker Equation. OGD regards online prediction as a linear regression problem. The stochastic gradient descent algorithm (SGD) is applied to optimize the weight vector iteratively. KF is based on the state-space model \cite{hangos2001intelligent}, which assumes an underlying hidden state for the dynamic system. KF is a typical EM algorithm that has a prediction step and a measurement step to make predictions and update the estimation of the hidden state using MLE \cite{scholz1985maximum} repeatedly until the convergence. Aside from simply replacing the missing values with predictions, multiple imputation \cite{Honaker2010What} is also an arising trend. This method basically imputes the missing values for multiple times. For each time, the filled sequence yields an estimation of the model parameters. The final result can be derived from the combination of all estimations. In statistics, unknown values are usually estimated by maximizing likelihood as parameters \cite{Jones1980Maximum}.

There are also other more subtle ways for imputation if the online setting is not required. A classic method is to employ Self-Organizing Maps (SOM) \cite{Kohonen1997Self, wang2003application, cottrell2007missing}, whose idea is to group similar fragments of the time series. Then the missing values in some time series fragments can be estimated using the clustering centers. Another more deterministic approach, Empirical Orthogonal Function (EOF) \cite{Preisendorfer1988Principal, Boyd1994Estimation}, takes the power of Singular Value Decomposition (SVD). By repeatedly decomposing the data matrix and reconstructing it with the $q$ largest singular values, the reconstructed results of the missing values can converge to give the best estimation. To make it even more accurate, a combination of SOM and EOF~\cite{Sorjamaa2007Time} was also proposed, which basically uses SOM to produce a good initialization for EOF. In addition to the semplice imputation strategy, we can also incorporate expert knowledge from the target area \cite{Honaker2010What}, which sometimes helps significantly. In order to improve the imputation performance, a recently proposed RNN-based method \cite{Che2016Recurrent} tries to exploit the pattern of missing values with the great representation ability of deep neural networks.

Another class of methods makes predictions directly rather than filling the missing values in advance. Based on OGD and the autoregressive (AR) model \cite{akaike1969fitting}, a missing value tolerant method for online time series prediction was proposed by Anava et al. \cite{anava2015online}. Its core idea is to dynamically extend the data in the current prediction window (with $d$ observations) according to the structure of the missing values. In order to cover all the possible cases, a weight vector with $2^d$ dimensions was constructed. The principle advantage compared with other methods is that it does not assume any structure on the missing data nor on the mechanism that generates the time series. In certain situations, it outperforms the original imputation-based OGD. However, it fails to give a satisfactory practical performance due to its exponential expand on the dimension of the weight vector. Such a problem of observation prediction with incomplete data can be also considered as classification of data with absent features. For instance, \cite{Shang2010Incomplete} employs the optimal hyperplane of max-margin classification to predict the future values. As we will disscuss later in detail, Attribute Efficient Ridge Regression (AERR) is also a direct approach to make predictions with limited attributes.

Although there are various methods to tackle the problem of time series prediction with missing values, not many of them can be easily applied on our attempted scenario because we would like to focus on the online setting and the assumption of AR model over the time series data. In this case, only the first several observations are given at start. After that, in each time step, one new observation is uncovered after the prediction. During this process, the AR model is learned from the data and at the same time applied to make predictions of future observations.

In our study, we adapted four methods (YW, KF, OGD, AERR) to become both missing value tolerant and online workable based on the AR model. Generally speaking, each method has its own pros and cons. YW has a rigorous mathematical foundation to ensure the optimality of the solution, but it has relatively high requirement on the stationarity of the time series, and its computational cost is also considerable due to the matrix inversion operations in equation solving. OGD is simple and has no requirement for sequence properties, but is susceptible to drastic change of the observation value in a short period of time, and the convergence cannot be guaranteed because of the influence of the learning rate. KF has great adaptability to different time series, but assumes gaussian noise. AERR avoids the adverse effect of poor imputation values, but it is difficult to adjust parameters for ideal performance.

Since different methods have different theoretical support and focus, for a given scenario, which one should be applied? In this paper, we study this problem experimentally by comparing these methods and analyzing their prediction accuracy in various data settings. To sum up, the major contributions of this paper are summarized as follows.

\begin{list}{\labelitemi}{\leftmargin=1em}\itemsep 0pt \parskip 0pt
\item We stress the importance of the problem of online prediction with missing values and for the first time present a systematic analysis and comparison of the various approaches tackling this problem experimentally.
\vspace{2mm}
\item We focused on four online methods with an offline method provided as the baseline. We conduct extensive experiments to comprehensively compare all methods from various aspects on both synthetic and real data.
\vspace{2mm}
\item We deeply analyze the experimental results and draw some noteworthy conclusions, which can contribute to industrial practice and provide some inspiration for researchers of related fields.
\end{list}

This paper is organized as follows. We elaborate the AR model and give an overview of our chosen methods in Section 2. We give detailed discussions of these methods respectively in Section 3. In Section 4, we illustrate our experimental setups and experimentally evaluate performance on prediction accuracy of each method and analyze their strengths and limitations in light of the comparison results. Last, we draw conclusions in Section 5.

\section{OVERVIEW}
As a fundamental model for time series analysis, AR model \cite{akaike1969fitting} is the most well-studied time series model. It and its deuterogenic models, such as ARMA (autoregressive moving average) \cite{Jones1980Maximum,Anava2013Online}, ARCH (autoregressive conditional heteroscedasticity) \cite{kirchgassner2013autoregressive}, ARIMA (autoregressive integrated moving average) \cite{ojo2010autoregressive}, have been used for forecasting in many fields like economics, medical science, climate analysis and so on \cite{earnest2005using}. It has been proved by the practice that such an assumption on the time series is reasonable and representative. Therefore, all our experiments are based on the AR model.

The idea of the AR model is quite simple. It assumes that each observation in the time series can be represented as a noisy and shifted linear combination of its past several observations. The AR coefficient vector $\bm{\alpha} \in \mathds{R}^p$, shifting constant $c$ and the noise $\epsilon$ are the three parameters that fully determine the characteristics of an AR process. With $X_t$ denoting the observation at time $t$, the mathematical form of an p-order AR model can be shown as follows:
\[X_t = c + \sum_{i=1}^p \bm{\alpha}^{(i)} X_{t-i} + \epsilon_t\]
where $\epsilon_t$ is the noise term at time $t$. In order to make a prediction of the current observation $X_t$, a number of the past observations and the parameters of the AR process are required. For simplicity, we only consider the zero-mean and stationary situation where the shifting constant $c=0$, and the coefficient vector $\bm{\alpha}$ does not change over time.
Our goal is to estimate the underlying $\bm{\alpha}$ by looking into the time series data, and then the prediction of the current observation is exactly the expectation of the weighted average of the past $p$ observations, which is
\[\tilde{X}_t=\sum_{i=1}^p \bm{\alpha}^{(i)} X_{t-i}\]
To this end, a number of methods have been developed. In this paper, we focus on four methods (YW, KF, OGD, AERR) that can be applied to perform online predictions of time series based on the AR model. An offline method, autoregressive least square (ARLS), is presented as an auxiliary tool to provide comparison baseline. We selected YW, KF and OGD, because they are classical and widely used. As for AERR, since it is based on multiple sampling and does not need missing value imputation for these parameters, it provides a different way to estimate the AR coefficients out of incomplete data, which is noteworthy. Here we briefly described all the five methods as follows.

\vspace{2mm}

\noindent \textbf{\emph{Yule-Walker Estimator} (YW)} is based on the famous Yule-Walker equation \cite{Zhang1992On,Weitian2011Solutions,Hallin1983Nonstationary}. There are two main operations in this method. One is the estimation of autocorrelation coefficients and the other is solving the equation to get an estimation of the AR model coefficients.

\vspace{2mm}
\noindent \textbf{\emph{Kalman Filter} (KF)} takes the advantage of the state-space model, which assumes a hidden state of a dynamic system. The hidden state contains the core information of the whole system, and observations are controlled by the hidden state but also influenced by the noise. KF is also an EM algorithm which makes predictions in the E step and updates the state estimation in the M step. Such two-step process is also known as prediction-measurement iteration.

\vspace{2mm}
\noindent \textbf{\emph{Online Gradient Descent} (OGD)} is actually a data-driven method that uses SGD to optimize the estimation of AR model coefficients. Its idea is to minimize the difference between last prediction and current observation by adaptively adjusting the weight. For such purpose, computing the gradient is the core operation of this method.

\vspace{2mm}
\noindent \textbf{\emph{Attribute Efficient Ridge Regression} (AERR)} is originally designed for the limited attribute observation (LAO) setting in linear regression problems. We adapted it to fit in our scenario of online prediction with missing values. Since it employs a sampling-based approach to compute the gradient, we can optimize the estimation of the AR coefficients without imputing the missing values.

\vspace{2mm}
\noindent \textbf{\emph{Autoregressive Least Square impute} (ARLS)}  \cite{Choong2009Autoregressive} is an offline method that is proposed to process missing data in the DNA micro array. Like Kalman Filter, this is also an EM-based method, which in each iteration, applies a least square estimator to compute AR coefficients and update missing values based on the idea of MLE. We select this method for providing baselines for other online methods.

\section{METHODS}
In this section, we describe the five selected methods in detail. Before the description, we would like to restate the scenario of the problem that all these methods are tackling. Our goal is to perform online prediction of time series with missing values. We assume that the time series is generated by a zero-mean $AR(p)$ model which basically means that each observation in the time series is a noise-involved linear combination of its former $p$ observations, where $p$ is the order of the model. We believe that such assumption is reasonable since the AR model has been widely used in the prediction of economics, informatics and natural phenomena \cite{akaike1969fitting}.

\subsection{Yule-Walker Equation}
The Yule-Walker Equation is the earliest approach to estimate parameters for a stationary time series with the linear prediction model. With the information given by autocorrelation coefficients, the underlying AR model coefficients can be easily derived by solving the equation.

To compute autocorrelation coefficients, we use second moment estimation. As for solving the equation, the only work is to compute the inversion of the autocorrelation matrix. A brief description of this algorithm is shown in \emph{Algorithm}~\ref{alg:YW}.

As is exhibited in \emph{Algorithm}~\ref{alg:YW}, first we have the time series set $\bm{y}_t,t\in\{1,\cdots,T\}$ and the order of the AR model $p$ as the input. Since we are under online setting, in each iteration, we only obtain one new observation, which may be a missing value. At beginning, we make sure that there is no missing values in the first $p$ observations and then initialize the AR model coefficients $\tilde{\bm{\alpha}}$ randomly. In each iteration, we first make prediction of the next observation using current $\tilde{\bm{\alpha}}$ and the last $p$ observations. Then we uncover the next observation $\bm{y}_t$ and adopt operations depending on its situation. If $\bm{y}_t$ is a missing value, we impute it with the last prediction value so that the current exposed sequence is complete. Then we update $\tilde{\bm{\alpha}}$ by recalculating the autocorrelation coefficients over this sequence and solving the equation. Otherwise, we perform the same operations without imputation.

\begin{algorithm}[H]
	\caption{Missing-value-tolerant YW}
	\label{alg:YW}
	\begin{algorithmic}
		\REQUIRE observation sequence $\{y_t\}_{t=1}^{t=T}$, AR model order $p$
		\ENSURE prediction sequence $\{ \tilde{y}_t \}_{t=p+1}^{t=T}$
		\STATE
		Initialize $\tilde{\bm{\alpha}}$ randomly
		\FOR{$t=p+1$ \TO $T$}
		\STATE{$\tilde{y}_t \gets \sum_{i=1}^{p} \tilde{\bm{\alpha}}^{(i)}y_{t-i}$}
		\IF{$y_t$ is missing}
		\STATE{Impute $y_t$ with $\tilde{y}_t$}
		\ENDIF
		\STATE{$\hat{\mu}_t \gets \frac{1}{t}\sum_{i=1}^{t}y_i$}						\STATE{$\hat{\sigma}_t \gets \frac{1}{t-1}\sum_{i=1}^{t}\left(y_i-\hat{\mu}_t\right)^2$}
		\STATE{$\gamma_0 \gets 1$}
		\FOR{$k=1$ \TO $p$}
		\STATE{
			$\gamma_k \gets \frac{1}{\hat{\sigma}_t(t-k)}\sum_{q=1}^{t-k}(y_q-\hat{\mu}_t)(y_{q+k}-\hat{\mu}_t)$
		}
		\ENDFOR
		\FOR{$i=1$ \TO $p$}
		\FOR{$j=1$ \TO $p$}
		\STATE{
			$\bm{R}^{(i,j)} \gets \gamma_{j-i}$\\
			$\bm{R}^{(j,i)} \gets \gamma_{j-i}$
		}
		\ENDFOR
		\ENDFOR
		\STATE{$\bm{\gamma} \gets [ \gamma_1, \gamma_2, \cdots, \gamma_p ] ^T$}
		\STATE{$\tilde{\bm{\alpha}} \gets \bm{R}^{-1} \bm{\gamma}$}
		\ENDFOR
	\end{algorithmic}
\end{algorithm}

\subsection{Kalman Filter}
Another influential method is Kalman Filter, which aims at exact Bayesain filtering for linear-Gaussian state space models. The state space model proposed along with the Kalman Filter in 1960s \cite{kalman1960new} has been one of the most important model to deal with the problem of time series prediction. Basically, a state space model (SSM) is very similar to an HMM while the difference is that its hidden states are continuous. In this case, a hidden state of a dynamic system is defined as the minimum information that contains all influence imposed by past external input and is sufficient to fully describe the future behavior of the system \cite{haykin2009neural}. In order to estimate the hidden state, a system model (or transition model) and an observation model were proposed to describe the process of a dynamic system \cite{Welch1995An}. In our scenario, the hidden state refers to the AR model coefficients. Thus, the system model characterize the variation of the AR model as time goes by. On the other hand, the observation model builds the relationship between the hidden state and the actual observations that is exposed to us. Originally, the hidden state of the state space model can change over time since it is influenced by its former state, external input and the system noise. However, since our problem is restricted to a zero-mean and stationary AR process, we assume that the hidden state remains the same. Likewise, in the observation model, we assume there is no external input so that the observation only depends on the current hidden state and the observation noise.

With all these restrictions and assumptions, the state space model for time series online prediction is described as below:
\[\bm{x}_t=\bm{x}_{t-1}\]
\[y_t=\bm{H}_t\bm{x}_t+\upsilon_t\]
\[\bm{x}_t=[\alpha_1,\alpha_2,\cdots,\alpha_p]^T\]
\[\bm{H}_t=[y_{t-1},y_{t-2},\cdots,y_{t-p}]\]
\[\bm{\upsilon}_t \sim N(0,\sigma^2)\]
Then we apply the Kalman filter to this model to estimate the hidden state and make prediction about future observations as the data stream in. A brief flow of this method is shown in \emph{Algorithm}~\ref{alg:KF}.

Let $\tilde{\bm{x}}_{t|\tau}$ denote $E(\bm{x}_t|y_{1:\tau})$. Basically, in each iteration, the learning strategy of KF is to first forecast $\tilde{\bm{x}}_{t|t-1}$ based on $\tilde{\bm{x}}_{t-1|t-1}$ and the transition model, and then update $\tilde{\bm{x}}_{t|t-1}$ to $\tilde{\bm{x}}_{t|t}$ with online maximum likelihood estimation (MLE) using the newly acquired observation. These two steps are called prediction step and measurement step, respectively. Actually, this is a typical \emph{EM} algorithm with the above two steps corresponding to \emph{E} step and \emph{M} step respectively.

To make it more specific, in the \emph{E} step, we derive the expectation of the hidden state $\tilde{\bm{x}}_{t|t-1}$ and its error covariance matrix $\bm{P}_{t|t-1}$ according to the transition model. Then we use $\tilde{\bm{x}}_{t|t-1}$ and $\bm{H}_t$ to make prediction of the next observation $\tilde{y}_t$. After that, we uncover the next observation. If it is a missing value, we impute it with the prediction value $\tilde{y}_t$. Then we come to the \emph{M} step where we update the Kalman gain $\bm{G}_t$ and compute the innovation $\bm{A}_t$. Based on these two values, we update the hidden state and error covariance matrix.

We should ensure that there is no missing values in the first $p$ observations because we need at least $p$ numbers for the very first prediction and then start the iteration. Besides, we can initialize $\tilde{\bm{x}}_{p|p}$ randomly and set $\bm{P}_{p|p}=c\bm{I}$ where $c$ represents a large positive integer, indicating that we have no prior knowledge about the hidden state.

\begin{algorithm}[t]
	\caption{Missing-value-tolerant KF}
	\label{alg:KF}
	\begin{algorithmic}
		\REQUIRE observation sequence $\{y_t\}_{t=1}^{t=T}$, AR model order $p$
		\ENSURE prediction sequence $\{\tilde{y}_t\}_{t=p+1}^{t=T}$
		\STATE
		Initialize $\tilde{\bm{x}}_{p|p}$ and $\bm{P}_{p|p}$
		\FOR{t=p+1 \TO T}
		\STATE{$\tilde{\bm{x}}_{t|t-1} \gets \tilde{\bm{x}}_{t-1|t-1}$}\\
		$\bm{P}_{t|t-1} \gets \bm{P}_{t-1|t-1}$\\
		\STATE{$\tilde{y}_t \gets \bm{H}_t \tilde{\bm{x}}_{t|t-1}$}
		\IF{$y_t$ is missing}
		\STATE{Impute $y_t$ with $\tilde{y}_t$}
		\ENDIF
		\STATE{$\bm{G}_t \gets \bm{P}_{t|t-1}\bm{H}_t^T\left(\bm{H}_t\bm{P}_{t|t-1}\bm{H}_t^T+\sigma^2\right)^{-1}$}\\
		$\bm{A}_t \gets y_t - \tilde{y}_t$\\
		$\tilde{\bm{x}}_{t|t} \gets \tilde{\bm{x}}_{t|t-1} + \bm{G}_t \bm{A}_t$\\
		$\bm{P}_{t|t} \gets \bm{P}_{t|t-1} - \bm{G}_t\bm{H}_t\bm{P}_{t|t-1}$
		\ENDFOR
	\end{algorithmic}
\end{algorithm}

\subsection{Online Gradient Descent}
Instead of regarding the time series as a linear dynamic system (LDS) with an estimable hidden state, OGD regards the estimation of AR coefficient as the problem of parameter optimization. It defines a loss function associated with the prediction accuracy. This is denoted by $l(X_t,\tilde{X}_t)$ measuring the error between the real observation $X_t$ and the prediction $\tilde{X}_t$ at time $t$. The most commonly used loss function is the least square loss as below:
\[l(X_t,\tilde{X}_t)=(X_t-\tilde{X}_t)^2\]
In our settings, the OGD Imputer adopts such form of loss function. The core idea of OGD is to use the gradient of the loss function to adjust AR model coefficients iteratively in order to minimize the loss. The training data conforms the following form:
\[\bm{x}_i=[X_{i+p-1},X_{i+p-2},\cdots,X_i]^T\]
\[y_i=X_{i+p}\]
With $\tilde{\bm{\alpha}}$ denoting the estimation of the AR model coefficients, the prediction of $X_t$ is described as follows:
\[\tilde{X}_t=\tilde{\bm{\alpha}} \bm{x}_{t-p}\]
Applying the least square loss function, we obtain
\[J(\tilde{\bm{\alpha}})=\frac{1}{m}\sum_{i=1}^{m}cost(\tilde{\bm{\alpha}},\bm{x}_i,y_i)\]
\[cost(\tilde{\bm{\alpha}},\bm{x}_i,y_i)=\frac{1}{2}\left(y_i-\tilde{\bm{\alpha}} \bm{x}_i\right)^2\]
OGD uses the gradient of $cost(\tilde{\bm{\alpha}},\bm{x}_i,y_i)$ to iteratively optimize $\tilde{\bm{\alpha}}$ as follows:
\[\tilde{\bm{\alpha}}_{t+1}=\tilde{\bm{\alpha}}_{t}-\eta\nabla cost(\tilde{\bm{\alpha}},\bm{x}_i,y_i)\]
where $\eta$ is the learning rate, and $t$ denotes the iteration time.

The flow of this method is shown in \emph{Algorithm}~\ref{alg:OGD}. We first randomly initialize the AR model coefficients and then repeatedly make prediction of the next observation based on current AR model coefficients and past $p$ observations. When we meet a missing value, we impute it with the last prediction value. At the end of each iteration, we update the AR coefficient by performing gradient descent on current observation loss.

By choosing different methods to initialize $\tilde{\bm{\alpha}}$, adjusting learning rate $\eta$ to an appropriate value or imposing additional constraints on $\tilde{\bm{\alpha}}$, we can improve the performance of online gradient descent. In this case, regularization techniques are commonly used to prevent overfitting. Two widely applied methods are \emph{Ridge} \cite{hoerl1970ridge} and \emph{Lasso} \cite{tibshirani1996regression}, which adds 2-norm and norm terms to the loss function. Since the proper use of regularization techniques matters a lot in some real applications, OGD only provides a basic structure where various adaptation can be conducted and incorporated with the original one. For instance, the \emph{AERR} method, as mentioned before, takes the advantage of \emph{Ridge} approach while updating the weight vector.

\begin{algorithm}[t]
	\caption{Missing-value-tolerant OGD}
	\label{alg:OGD}
	\begin{algorithmic}
		\REQUIRE observation sequence $\{X_t\}_{t=1}^{t=T}$, learning rate $\eta$, AR model order $p$
		\ENSURE prediction sequence $\{\tilde{X}_t\}_{t=p+1}^{t=T}$
		\STATE{Initialize $\tilde{\bm{\alpha}}$ randomly}
		\FOR{t=p+1 \TO T}
		\STATE{$\tilde{X}_{t} \gets \tilde{\bm{\alpha}} \bm{x}_{t-p}$}
		\IF{$X_t$ is missing}
		\STATE{Impute $X_t$ with $\tilde{X}_t$}
		\ENDIF
		\STATE{$\tilde{\bm{\alpha}} \gets \tilde{\bm{\alpha}}-\eta\nabla cost(\tilde{\bm{\alpha}},\bm{x}_{t-p},y_{t-p})$}
		\ENDFOR
	\end{algorithmic}
\end{algorithm}

\subsection{Attribute Efficient Ridge Regression}
Some researchers think that simply using predictions to impute missing values will slow down the convergence of algorithms and even mislead later predictions. Instead of imputation, some other mechanisms are proposed for dealing with missing observations. \cite{W1981Estimation} makes assumptions of the generating mechanism of the missing values. \cite{Ding2010Time} considers a scarce pattern of missing values in the observed signals. \cite{Anava2015Online2} extends the weight vector's dimensions in order to learn from the structure of missing data. Although they all claim to be ideal, the different restrictions on data and the unsatisfactory time complexity makes them specious. \cite{W1981Estimation} and \cite{Ding2010Time} are both with strict restrictions on data, and they are not under online settings, which makes it difficult to compare them with other methods experimentally. Although the method in \cite{anava2015online} is originally under our online settings and has simple assumptions, the extension in the algorithm makes it too slow to be included in our comparisons.

Since the imputation-absent methods that are completely applicable to our scenario are rare and their limitations and restrictions make it difficult to design experiments, we adapt AERR \cite{Hazan2012Linear} as an online semi-imputation time series prediction methods and include it to our experiments as a trial, where semi-imputation means that the previous prediction results are only used for the future predictions, and the update of the estimated AR coefficients only depends the existing data. A brief process of this method is shown in \emph{Algorithm} \ref{alg:AERR}.

According to \emph{Algorithm} \ref{alg:AERR}, the main operation of AERR is a sampling to compute the gradient. In each iteration, there are two sampling steps. Firstly, we sample from the past $p$ observations uniformly for $k$ times to obtain an estimation of their expectation. Secondly, we choose from the past $p$ observations for another time. In this sampling step, each probability of observation selection is associated with the current \boldmath $\frac{(a_{l}[j])^2}{||a_{l}||^2}$. Then we multiply these two estimations to get an unbiased estimation of the gradient. If we unfortunately get a missing value during the sampling process, we discard this sample result.

We use filled sequence $\bm{\bar{y}}$ to produce prediction $P$, so that we can fully use current AR coefficients $\bm{\alpha}$ to obtain predictions. Meanwhile we sample from sequence $\bm{\tilde{y}}$, which has missing observations represented by $*$, in order to prevent early poor predictions from continuously misleading the later update of the AR coefficients. Assume that the missing probability of each observation is equal, the missing rate $\gamma$ is used to make up for missing values and obtain unbiased estimations \cite{DBLP:conf/bigdata/WeiWSGL17}.

As described above, with slight adjustment for our intended case, we can add it to our experiments to verify how well the strategy of sampling may preform in online time series prediction with missing values.

\begin{algorithm}[htbp]
	\caption{Missing-value-tolerant AERR}
	\label{alg:AERR}
	Parameters: $B>0,\eta>0,p>0$
	\begin{algorithmic}
		\REQUIRE observation sequence $\{\bm{y}_t\}_{t=1}^{t=T}$, missing rate $\gamma$, sample time $k$
		\ENSURE predictions $\{P_t\}_{t=p+1}^{t=T}$
		\STATE
		Initialize $\bm{\alpha}_0$
		\STATE $\bm{\bar{y}}_{1:p} \gets \bm{y_{1:p}}$, $\bm{\tilde{y}}_{1:p} \gets y_{1:p}$\\
		\STATE $l \gets 0$
		\FOR{t=1 \TO T-p}
		\STATE $\bm{\bar{x}}_t \gets \bm{\bar{y}}_{t:t+p-1}$
		\STATE $\bm{\tilde{x}}_t \gets \bm{\tilde{y}}_{t:t+p-1}$
		\STATE $\bm{\tilde{\alpha}}_t \gets \frac{1}{l+1} \sum_{0}^{l} \bm{\alpha}_l $
		\STATE $P_{t+p} \gets \bm{\tilde{\alpha}}_t \bm{\bar{x}}_t$
		\IF{$\bm{y}_{t+p}$ is missing}
		\STATE $\tilde{\bm{y}}_{t+p} \gets *$, $\bm{\bar{y}}_{t+p} \gets P_{t+p}$
		\STATE continue
		\ELSE
		\STATE $\tilde{\bm{y}}_{t+p} \gets \bm{y}_{t+p}, \bm{\bar{y}}_{t+p} \gets \bm{y}_{t+p}$
		\ENDIF
		\FOR{j=1 \TO k}
		\STATE Pick $i_{j,t} \in [p]$ uniformly
		\IF{$\bm{\tilde{x}}_t[i_{j, t}] \neq *$}
		\STATE ${\bm{x}}_{j, t} \gets p\bm{\tilde{x}}_t[i_{j,t}] \bm{e}_{i_{j,t}}/(1-\gamma)$
		\ENDIF
		\ENDFOR
		\STATE $\bm{x}_t \gets \frac{1}{k}  \sum_{j=1}^{k}\bm{x}_{j,t}$
		\STATE Choose $j$ with probability $\frac{\bm{\alpha}_l[j]^2}{||\bm{\alpha}_l||^2}$
		\IF{$\bm{\tilde{x}}_t[j] \neq *$}
		\STATE $\bm{\phi} \gets \frac{||\bm{\alpha}_l||^2\bm{\tilde{x}_t[j]}}{\bm{\alpha}_l[j](1-\gamma)}-y_{t+p}$
		\ELSE
		\STATE $\phi_t \gets 0$
		\ENDIF
		\STATE $\bm{g}_t \gets \phi_t \bm{x}_t $
		\STATE $\bm{v}_t \gets \bm{\alpha}_l - \eta\bm{g}_t$
		\STATE $\bm{\alpha}_{l+1} \gets \frac{\bm{v}_t B}{max(||v_t||, B)}$
		\STATE $l \gets l+1$
		\ENDFOR
	\end{algorithmic}
\end{algorithm}

\subsection{Autoregressive Least Square}
ARLS is expected to achieve better prediction performance over other online methods because it is an iterative and offline method, which means that it can perform global optimization on the whole time series. The basic idea of this method is to estimate AR weights and recover the missing values at the same time through an iterative way. Initially, it sets all the missing values to zero. Then, in each iteration, it traverses all the observations and optimizes the AR parameter based on regression, after which all the missing values are imputed by doing inference with current AR weight and the known observations. When all the missing values and AR weights converge, we obtain the final result. A brief process of this method is shown in \emph{Algorithm}~\ref{alg:ARLS}.

\begin{algorithm}[htbp]
	\caption{ARLSimpute}
	\label{alg:ARLS}
	\begin{algorithmic}
		\REQUIRE observation sequence $\{y_t\}_{t=1}^{t=T}$, iteration time $N$, regression model \emph{$\bm{M}$}
		\ENSURE optimized AR parameters $\tilde{\bm{\alpha}}$, imputed observation sequence $\{\tilde{y}_t\}_{t=1}^{t=T}$
		\STATE
		Initialize $\tilde{\bm{\alpha}}$\\
		\FOR{t=1 \TO T}
		\IF{$y_t$ is missing}
		\STATE $\tilde{y}_t \gets 0$
		\ELSE
		\STATE $\tilde{y}_t \gets y_t$
		\ENDIF
		\ENDFOR
		\FOR{i=1 \TO N}
		\STATE optimize $\tilde{\bm{\alpha}}$ based on $\{\tilde{y}_t\}_{t=1}^{t=T}$ and \emph{$\bm{M}$}
		\FOR{t=1 \TO T}
		\IF{$\bm{y}_t$ is missing}
		\STATE $\tilde{y}_t \gets \tilde{\bm{\alpha}}\tilde{y}_{t-p:t-1}$
		\ENDIF
		\ENDFOR
		\ENDFOR
	\end{algorithmic}
\end{algorithm}

\section{EXPERIMENTAL EVALUATION}
In this section, we evaluate the selected methods experimentally. Note that though ARLS is not an online method, we also take it into our experiments in order to provide baselines for better understanding of how well each method performs. To quantitatively show the accuracy of the prediction results, there are normally several frequently used measurements in time series analysis, such as MSE (mean square error), RMSE (root mean square error), MAPE (mean absolute percentage error) and MAD(mean absolute difference). Since in our work, all the time series are normalized with zero mean, and there is no comparison of prediction accuracy across different time series, MSE is a simple but reasonable measure to make the assessment. Therefore, we use it as a measurement of the prediction accuracy. The mathematical definition of MSE is as follows:
\[MSE \left( y_{1:T},\tilde{y}_{1:T} \right)=\frac{1}{T}\sum_{t=1}^{T} \left( y_t-\tilde{y}_t \right) ^2\] where $y_i$ and $\tilde{y}_i$ denote the actual observation and prediction value at time $t$, respectively, and $T$ represents the length of the time series.

Firstly, we describe the design of the experiments. We have a uniform standard setting for all experiments. It gives a set of default values of parameters for generating the synthetic data and controlling the order of the prediction model. The data generation parameter is denoted by $\theta=(L,\sigma^2,\alpha,miss)$ with following default values:

\[L=2000, \sigma^2=0.3^2\]
\[\alpha=[0.3, -0.4, 0.4, -0.5, 0.6]\]
\[miss\in\{0,0.05,0.1,0.15,0.2,0.25,0.3\}\]

Note that for all experiments, the missing rate varies from 0 to 0.3 with a step of 0.05. They respectively represent the length of each time series, the variance of gaussian white noise in the AR model, the AR model coefficient vector, and the missing rates. The order of the prediction model is denoted by $p_{fit}$. By default, $p_{fit}=5$. On the basis of the standard setting, we conduct experiments on synthetic data and focus on one single parameter in each experiment and study the influence of this parameter on each method when its value varies in a given range.

For some experiments on synthetic data, we choose the parameter values according to the real time series data, trying to hold more practical significance. We find that the length of real time series from \emph{UCR Archive} and \emph{Tushare}, which are the data source of the later real data experiments, roughly ranges from 50 to 3000. We set a parameter $L\in\{1000,2000,3000,4000,5000\}$. Also we notice that the observation values of the normalized time series mostly fall in the range of -1.5 to 1.5, so a noise with the standard deviation of $\sigma\in\{0.3,0.6,0.9,1.2,1.5\}$ approximately takes 10 to 50 percent of the amplitude, which can be quite common in the low SNR environment. As for the AR coefficients, since not all coefficient vectors can generate a stationary time series, we try to find some suitable cases by trying different permutations and combinations of numbers from $\left\lbrace\frac{i-10}{10}|i\in[1,19],i\in\mathds{Z}\right\rbrace$.

At Last, we selected various types of real time series data with different length, local trend, seasonality, degree of fluctuation and smoothness for comparisons. By running the test in such complex environment, we try to show some interesting features of these methods concerning their ability to cope with certain types of time series.

\subsection{Impact of Missing Rates}
\label{sec:missingrate}
\begin{figure}[htbp]
	\centering
	\includegraphics[width=0.48\textwidth]{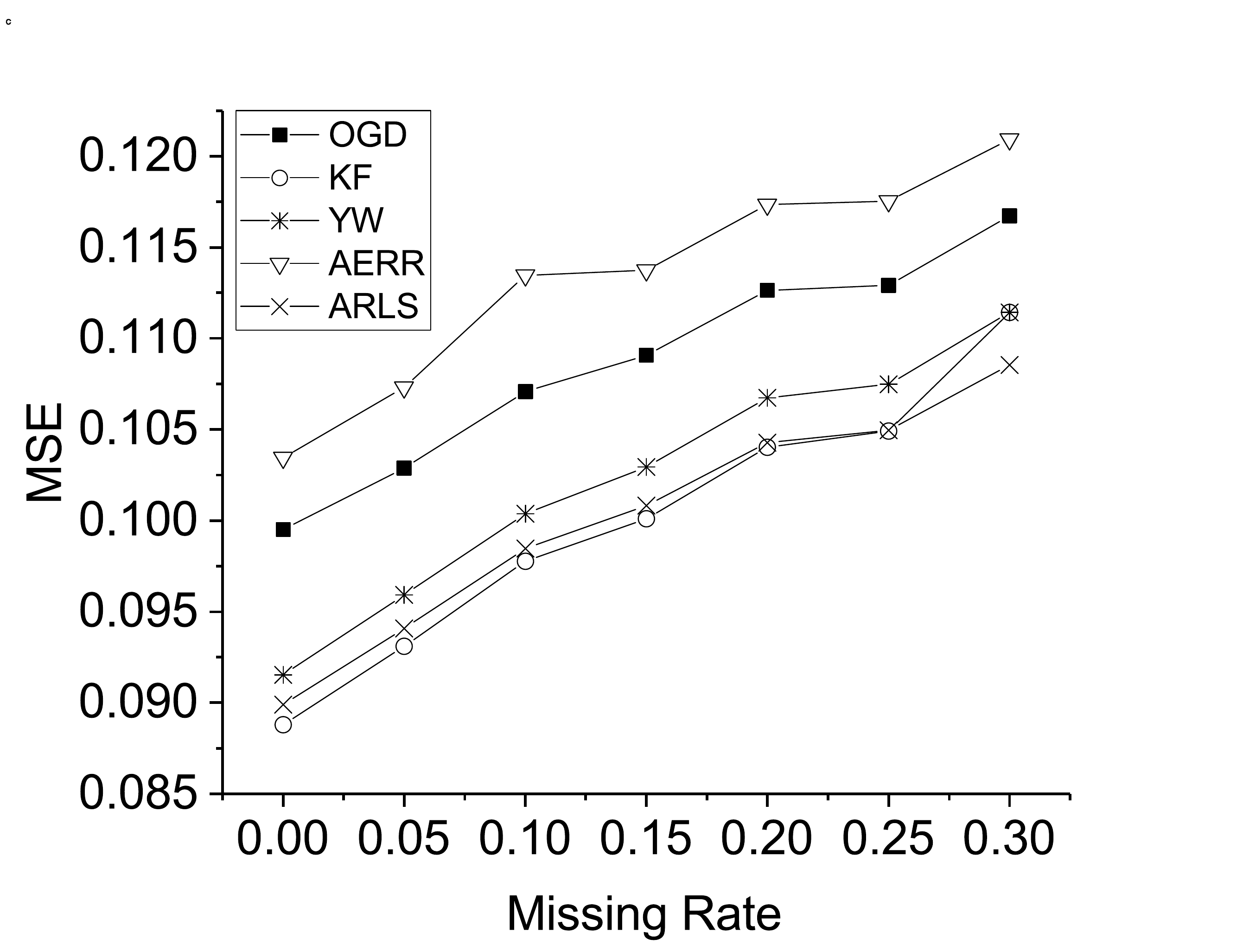}
	\caption{MSE VS. missing Rates}
	\label{fig:std1}
\end{figure}

In the first experiment, we compare methods on the prediction accuracy as the missing rates increases. For each setting, we generate 20 different time series with randomized initialization. Missing values are generated by setting each value to empty with a certain missing rate. The experimental results are shown in Figure \ref{fig:std1}.

As illustrated in the Figure \ref{fig:std1}, with the increase of the missing rate, the MSEs of all the methods have an approximately linear growth. Apparently, higher missing rates yield higher prediction errors, but the increase of the latter is fairly gentle. Although the growing trend of each method's MSE is similar, there is still discernible difference among these five methods.

As an online method, Kalman Filter (KF) performs considerably good. Its MSE is highly close to that of the offline method ARLS, and such advantage keeps on when the missing rate increases. As shown in the Figure \ref{fig:std1}, MSEs of both methods are close to the variance of the white noise. Since in maximum likelihood estimation (MLE), the ideal MSE equals to the variance of the white noise \cite{scholz1985maximum}, similarly, such results is nearly ideal. Such results are caused by the MLE in the measurement step in KF. The way that Kalman filter updates the posterior estimation of the hidden state is known as the recursive least square (RLS) algorithm \cite{grant1987recursive}. It performs step-size adaptation to the online time series. Despite the fact that the algorithm needs only one pass over the data, the historical information is stored in both the hidden state estimation and the Kalman Gain. As time goes by, the prior knowledge about the distribution of the hidden state is accumulated so that it can soon converge to the optimal posterior. The process of its adaptation to the observation sequence is similar as object tracking, which is the earliest application of Kalman filtering.

We also observe that the widely used YW has a quite similar performance with KF. Although it still has overall larger MSEs, the narrow gap does not expand as the missing rate goes up, and eventually they have roughly the same performance when reaching the highest missing rate. Due to the simplicity of YW and its close performance as the best KF, in certain cases, it is a better choice. However, when high efficiency is required, YW will probably not work because its time cost is too high. There are two main time expenses of YW. One is the time cost on the computation of autocorrelation coefficients (short in ACORR). Originally, as the time series grows, this cost also grows. However, we can easily reduce this cost to a constant by making good use of the previous computational results. The other one is the time cost on solving the equation. This operation needs to compute the inverse of the matrix, which is very time consuming.

Different from YW, OGD has a comparatively worse prediction results. Since this method employed stochastic gradient descent algorithm to optimize the AR coefficients, its performance is bound to a local training data. Owing to the online setting, in each iteration, only one observation in the whole data space is disclosed to the training model. Therefore, the new gradient is easy to be affected by the randomness of data due to the presence of the white noise. Furthermore, to ensure that the optimization process does not collapse, the learning rate cannot be too large, which has restricted the convergence speed of this method. On the other hand, AERR has the highest MSE under all missing rate settings. This has shown the strength of imputation strategy to some extent.

\subsection{Impact of Time Series Length}
\label{sec:tslength}
\begin{figure}[htbp]
	\centering
	\subfigure[$L=1000$]{
		\label{fig:ext1:1000}
		\includegraphics[width=0.48\textwidth]{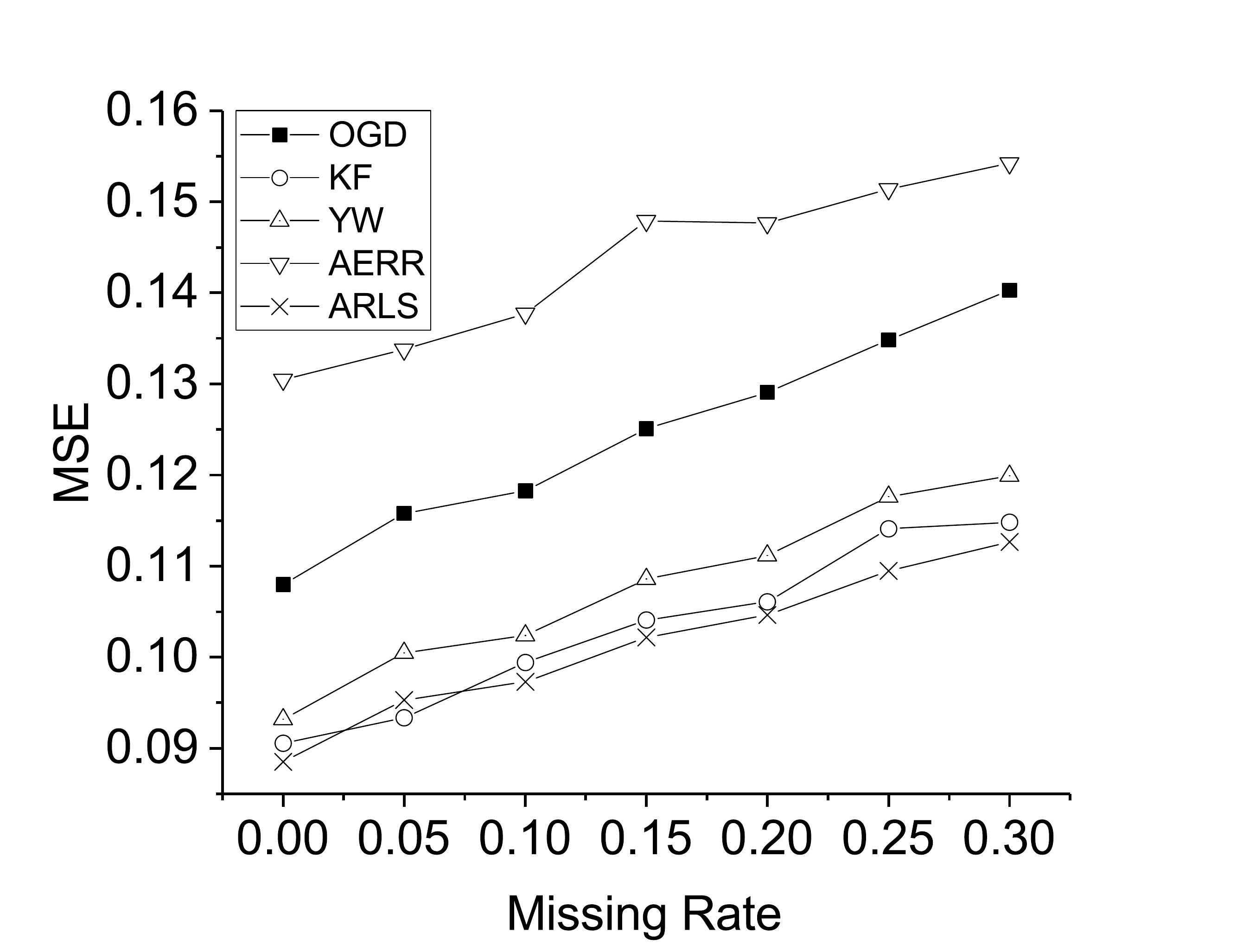}}
	\subfigure[$L=5000$]{
		\label{fig:ext1:5000}
		\includegraphics[width=0.48\textwidth]{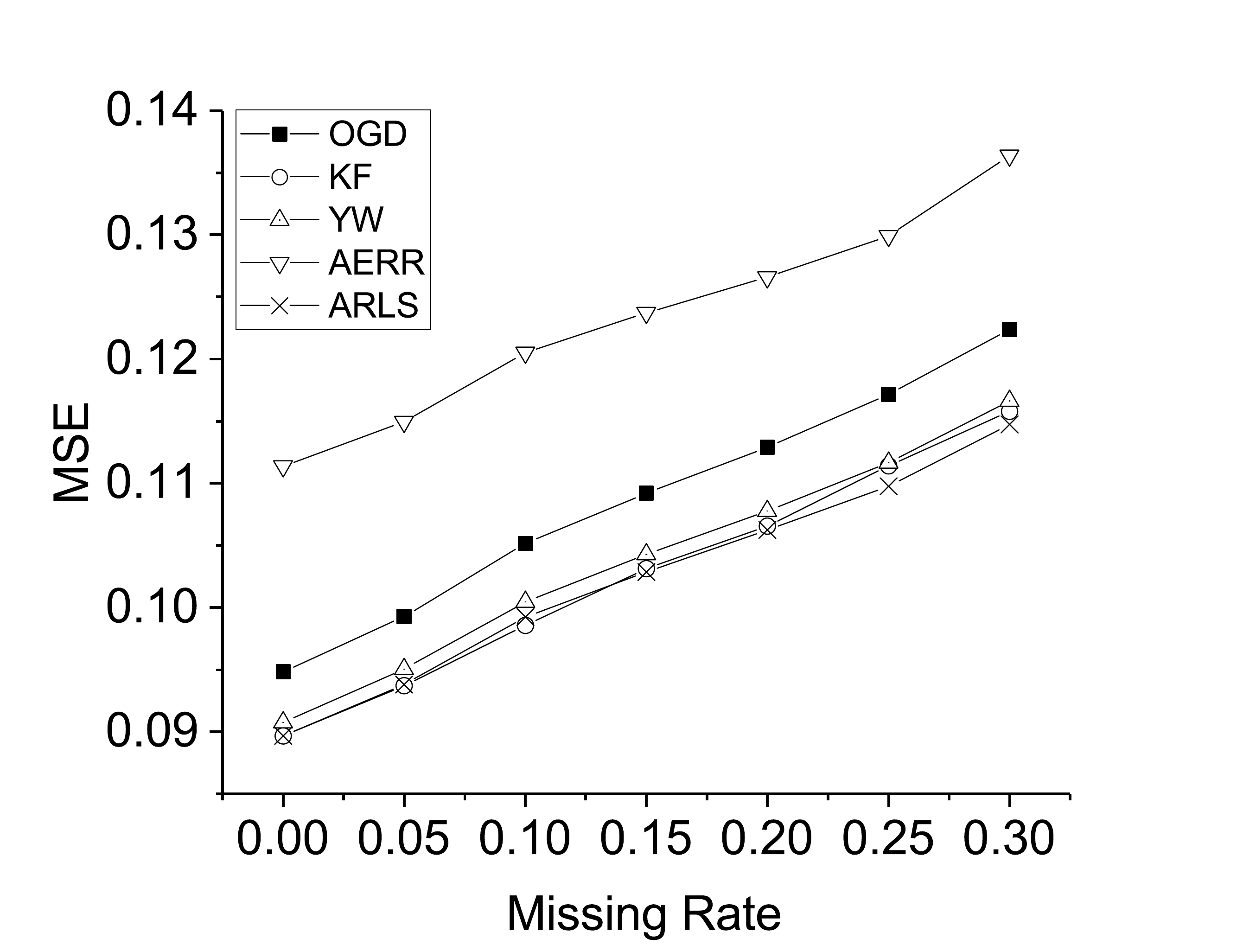}}
	\subfigure[$miss=0.05$]{
		\label{fig:ext1:0.05}
		\includegraphics[width=0.48\textwidth]{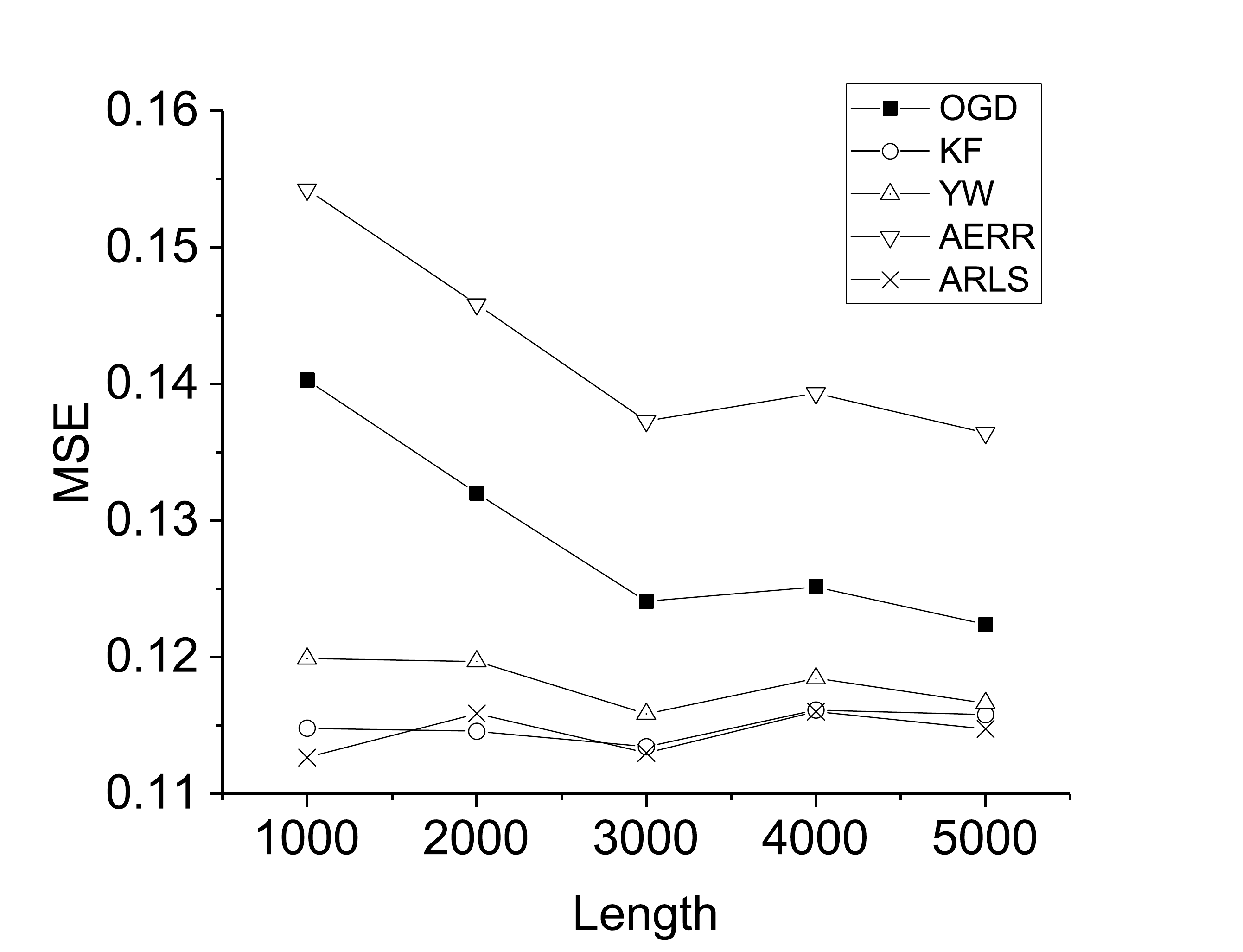}}
	\subfigure[$miss=0.3$]{
		\label{fig:ext1:0.3}
		\includegraphics[width=0.48\textwidth]{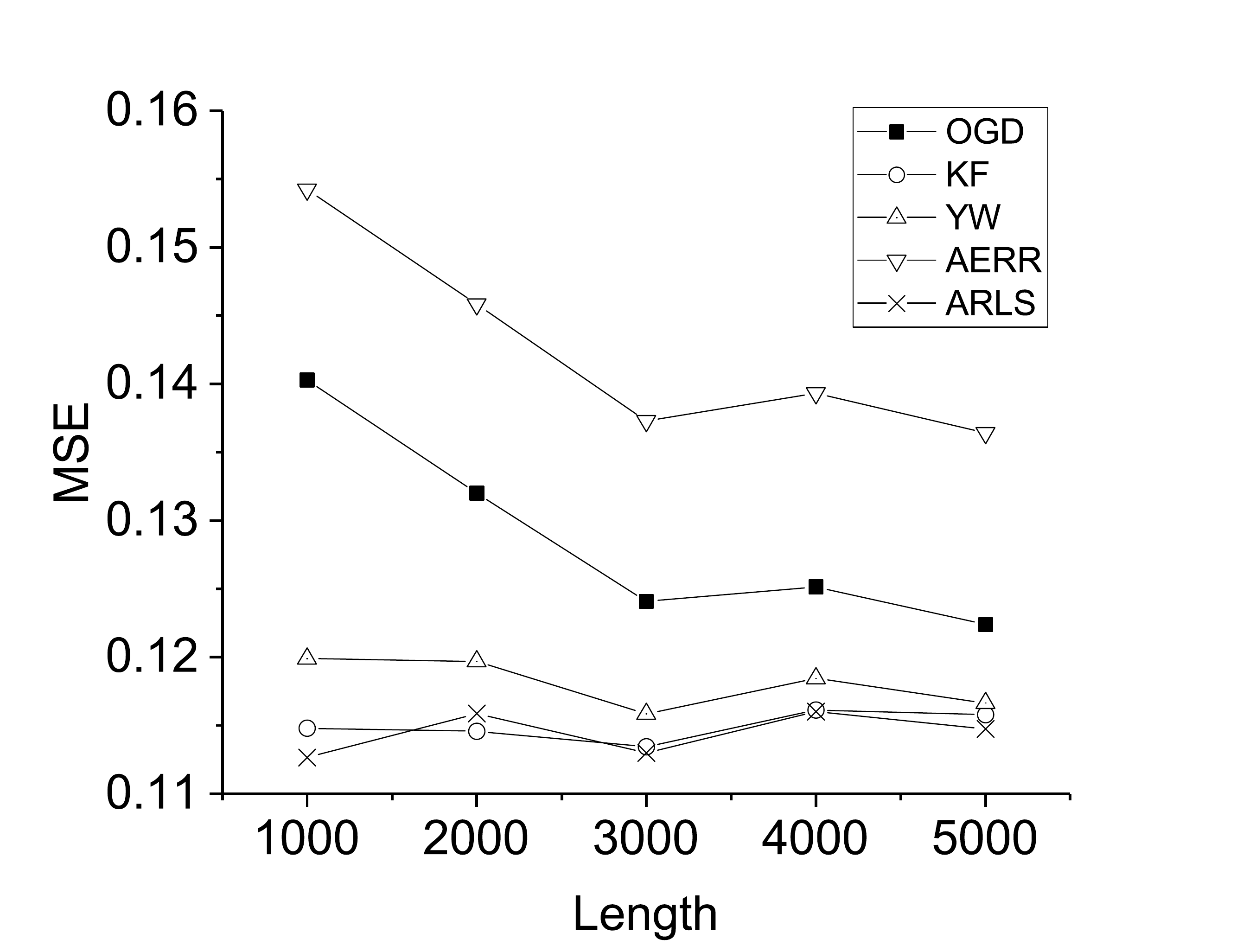}}
	\caption{MSE VS. Time Series Length}
	\label{fig:ext1}
\end{figure}

In this experiment, we adjust the length of the time series to test its impact on the method performance. On the basis of the standard setting, we vary the length from 1000 to 5000. For each length, we generate 20 time series as our input data for each length and missing rate. The experimental results are shown in Figure \ref{fig:ext1}.

It is observed from the experimental results that the MSE of all methods increases with the increase of the missing rate. However, their relative order does not change. This feature is consistent under five data lengths. With the increase of the time series length, the MSE of the OGD and AERR significantly decreases; the MSE of YW decreases slightly; and the MSE of the KF and ARLS methods does not change significantly. It is observed that although the relative order of the predictive performance of the various methods has not changed, the gap between them decreases gradually as the amount of training data increases.

This is caused by the limitations of the random gradient drop (SGD) algorithm. Both the AERR and OGD methods use the SGD strategy to optimize the parameters of the model. This optimization is first-order \cite[p.~41]{Nocedal2006NO}, and the convergence rate of the parameters is affected by the learning rate and the randomness of the current training samples. This leads to a slower convergence of the AERR and OGD method on the same time series. Such results suggest that it is better to have sufficient data when using OGD and AERR methods for time series prediction.

\subsection{Impact of Noise Variance}
\label{sec:noisevariance}
\begin{figure}[htbp]
	\centering
	\subfigure[$\sigma=0.3$]{
		\label{fig:ext2:0.3}
		\includegraphics[width=0.48\textwidth]{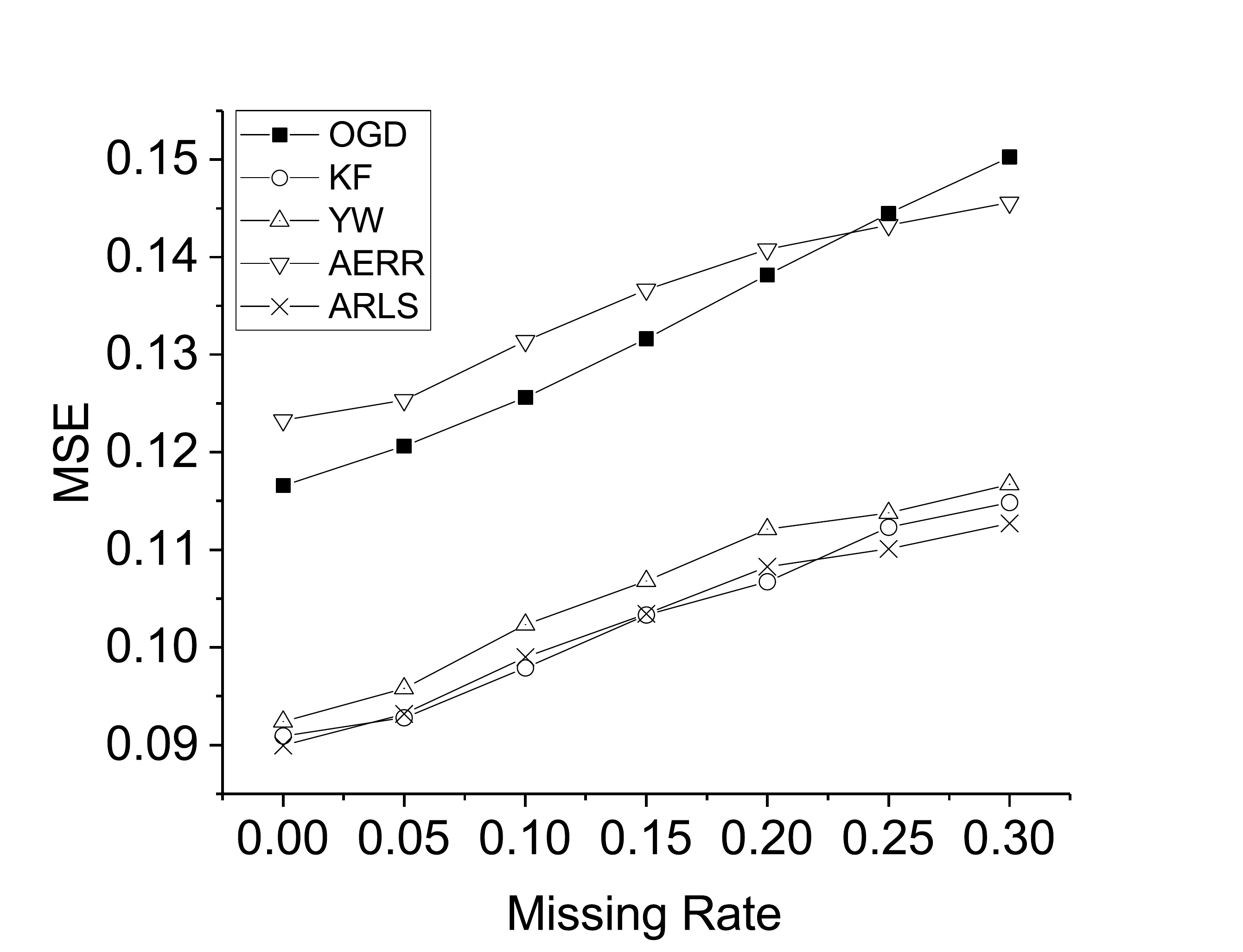}}
	\subfigure[$\sigma=1.5$]{
		\label{fig:ext2:1.5}
		\includegraphics[width=0.48\textwidth]{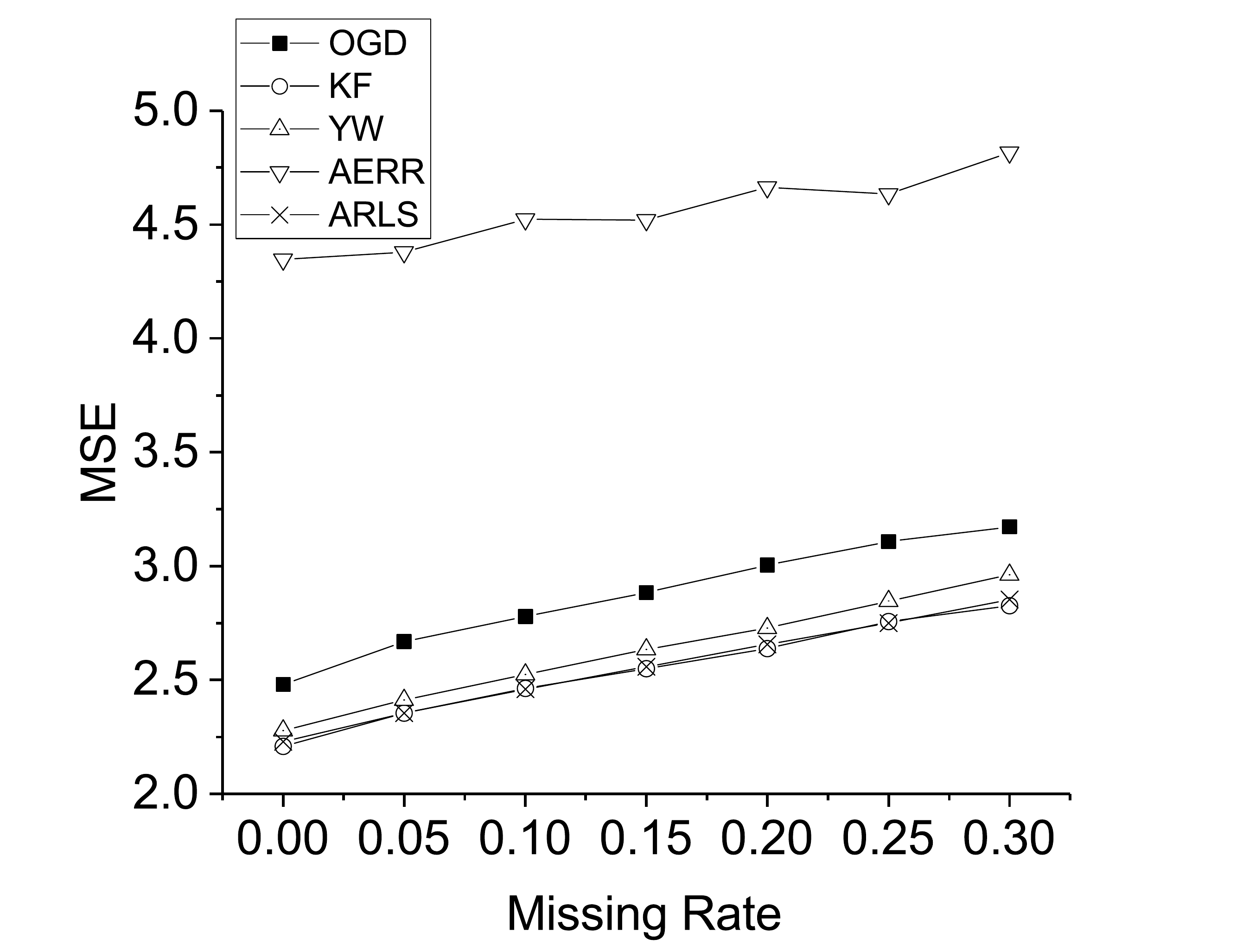}}
	\subfigure[$miss=0.05$]{
		\label{fig:ext2:0.05}
		\includegraphics[width=0.48\textwidth]{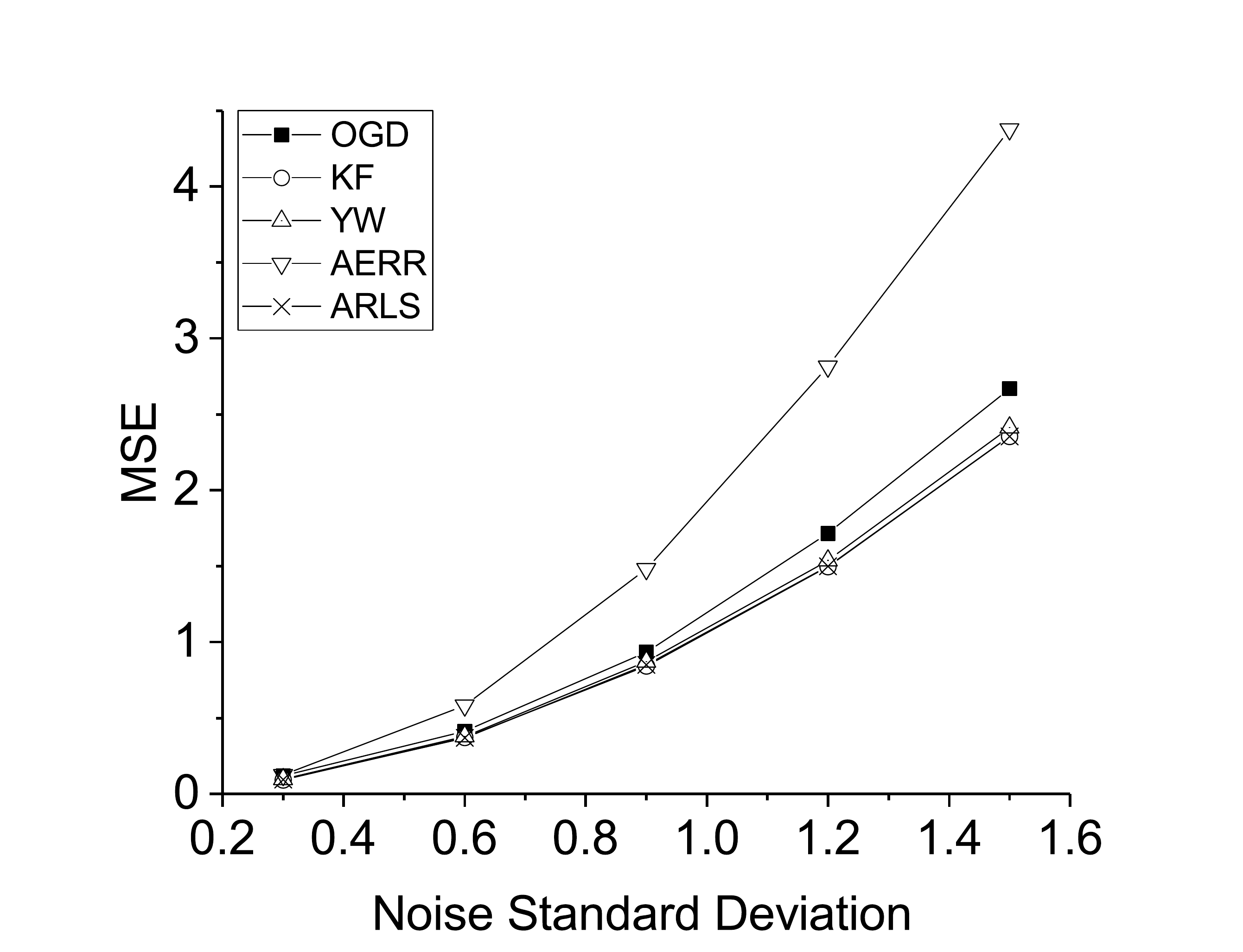}}
	\subfigure[$miss=0.3$]{
		\label{fig:ext2:0.3m}
		\includegraphics[width=0.48\textwidth]{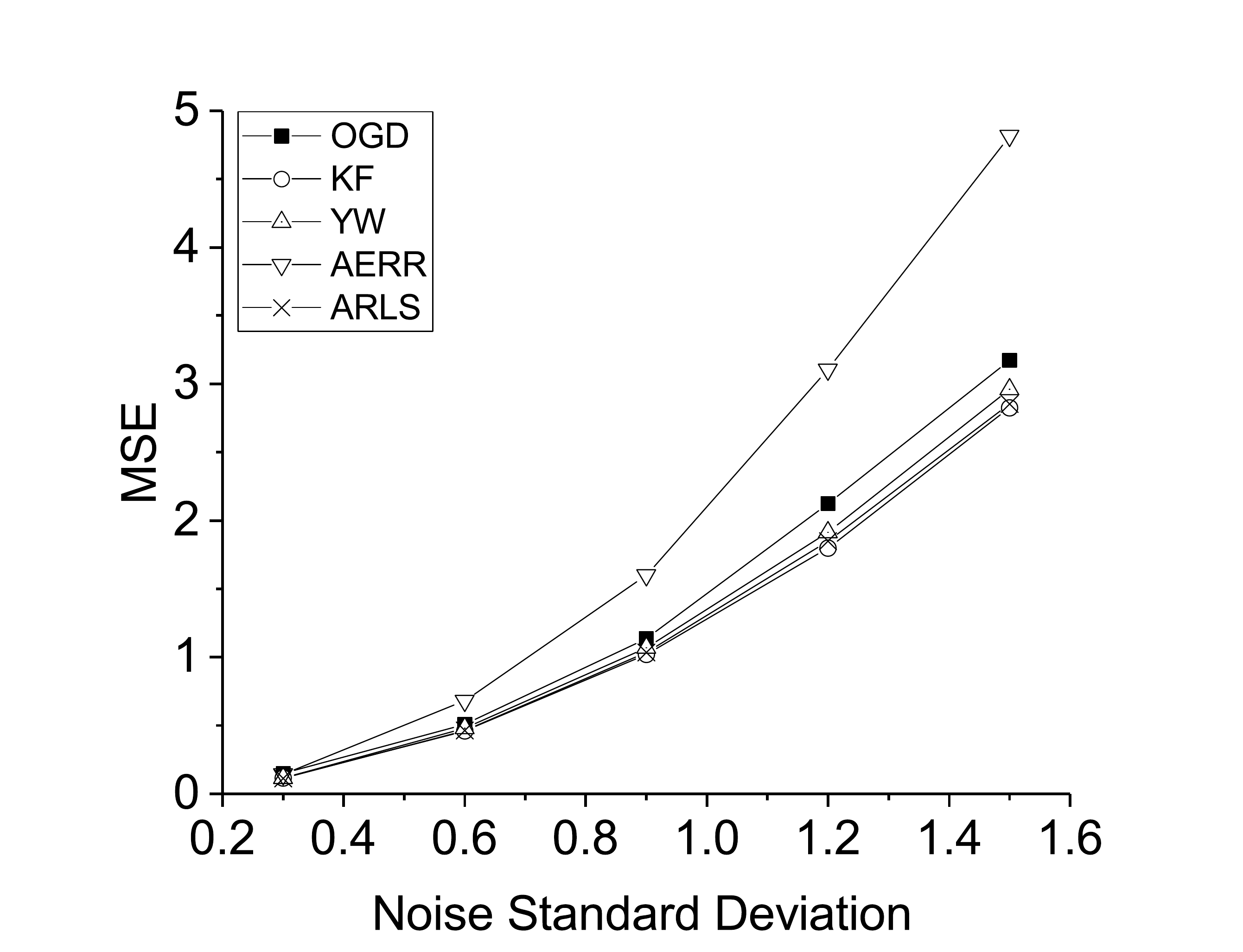}}
	\caption{MSE VS. Noise Variance}
	\label{fig:ext2}
\end{figure}

In this experiment, we vary the variance($\sigma$) of the white noise from 0.3 to 1.5. For each noise variance and each missing rate, we generate 20 time series. The experimental results are shown in Figure \ref{fig:ext2}.

Obviously, with the increase of the variance, the MSE of all methods increases. YW, KF and ARLS achieve very similar results. Compared with them, the MSE of OGD is larger, but the gap substantially remains stable as the variance increases. In contrast, the difference between the MSE of AERR and the best MSE is significantly expanded with the increase of variance.

There are two reasons for this result. First, the original design of AERR aims at any-time settings rather than online ones. In order to make AERR work in the online setting, we have to use the weight to make predictions before the average computation. Therefore, the randomness of the weight reduced the accuracy. Second, it is observed from the \emph{Algorithm}~\ref{alg:AERR} that the operation of weight updating depends on the sampling results of the past $p$ observations. According to \emph{Algorithm} \ref{alg:AERR}, when we happen to pick the missing observation in the sampling second step, the gradient is set to be zero so that the weight will not update. If the missing rate is high, this often happens, which will result in the low speed of weight convergence. Thus, the predictions produced by the not-converging weight are unable to track the rapid changes in the observation sequence. In the other words, AERR is comparatively dull to data changes. Therefore, when the noise variance of the time series gets larger, the gap between the predicted value and the observed one is widened because the sequence changes more dramatically over time.

\subsection{Impact of AR Model Coefficients}
\label{sec:armodelcoeff}
\begin{figure}[htbp]
	\centering
	\subfigure[Coefficients No.1]{
		\label{fig:ext3:1}
		\includegraphics[width=0.48\textwidth]{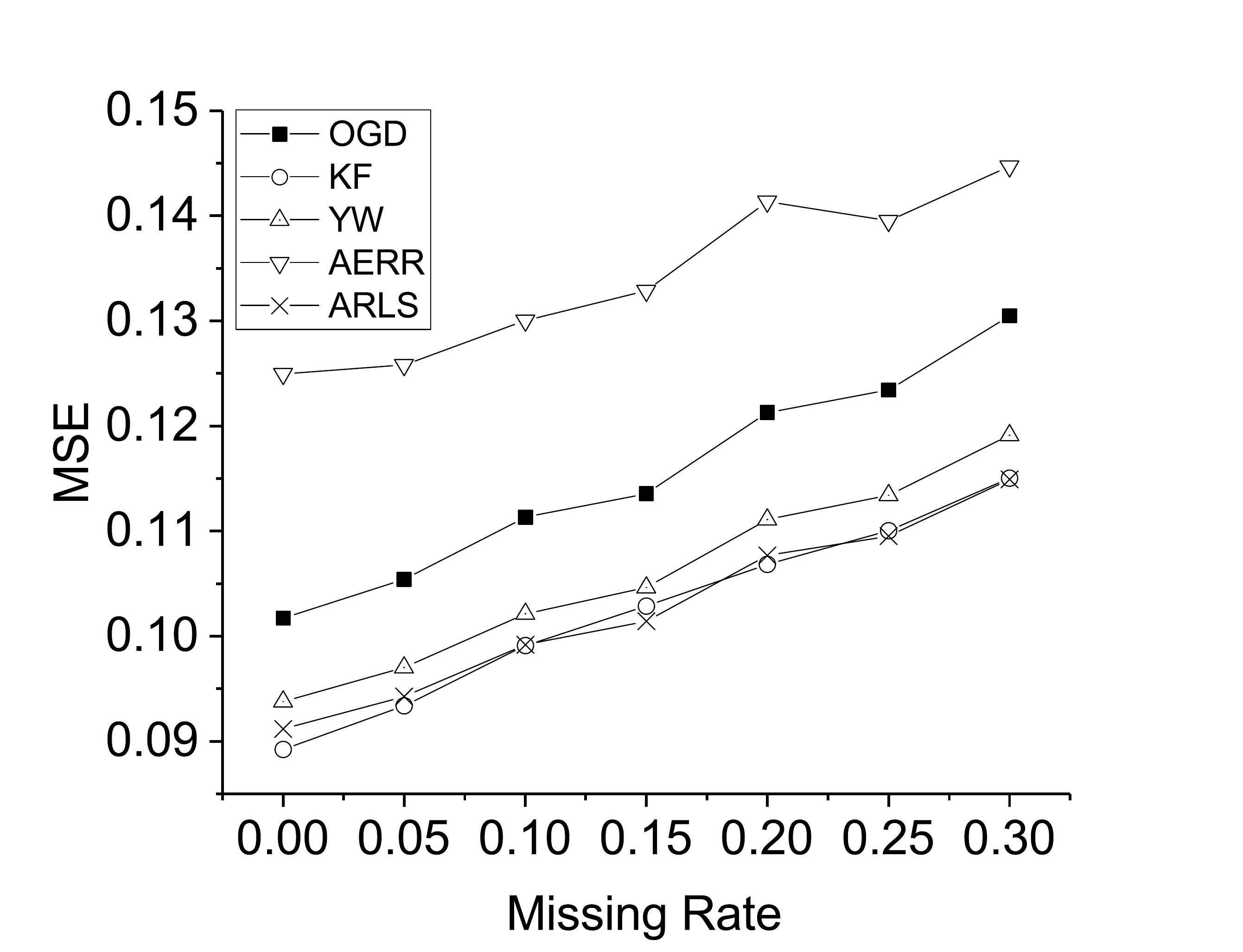}}
	\subfigure[Coefficients No.2]{
		\label{fig:ext3:2}
		\includegraphics[width=0.48\textwidth]{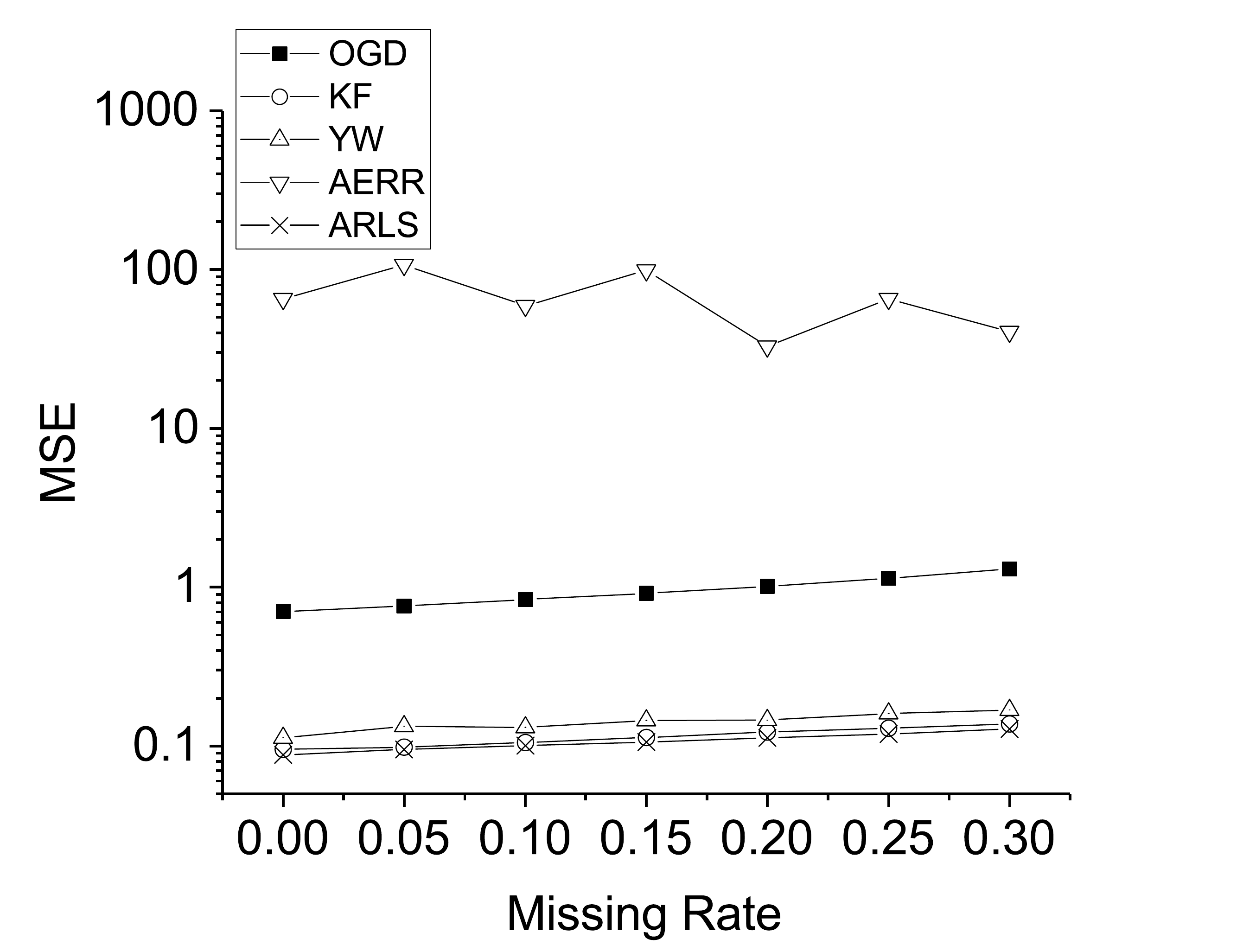}}
	\subfigure[Coefficients No.3]{
		\label{fig:ext3:3}
		\includegraphics[width=0.48\textwidth]{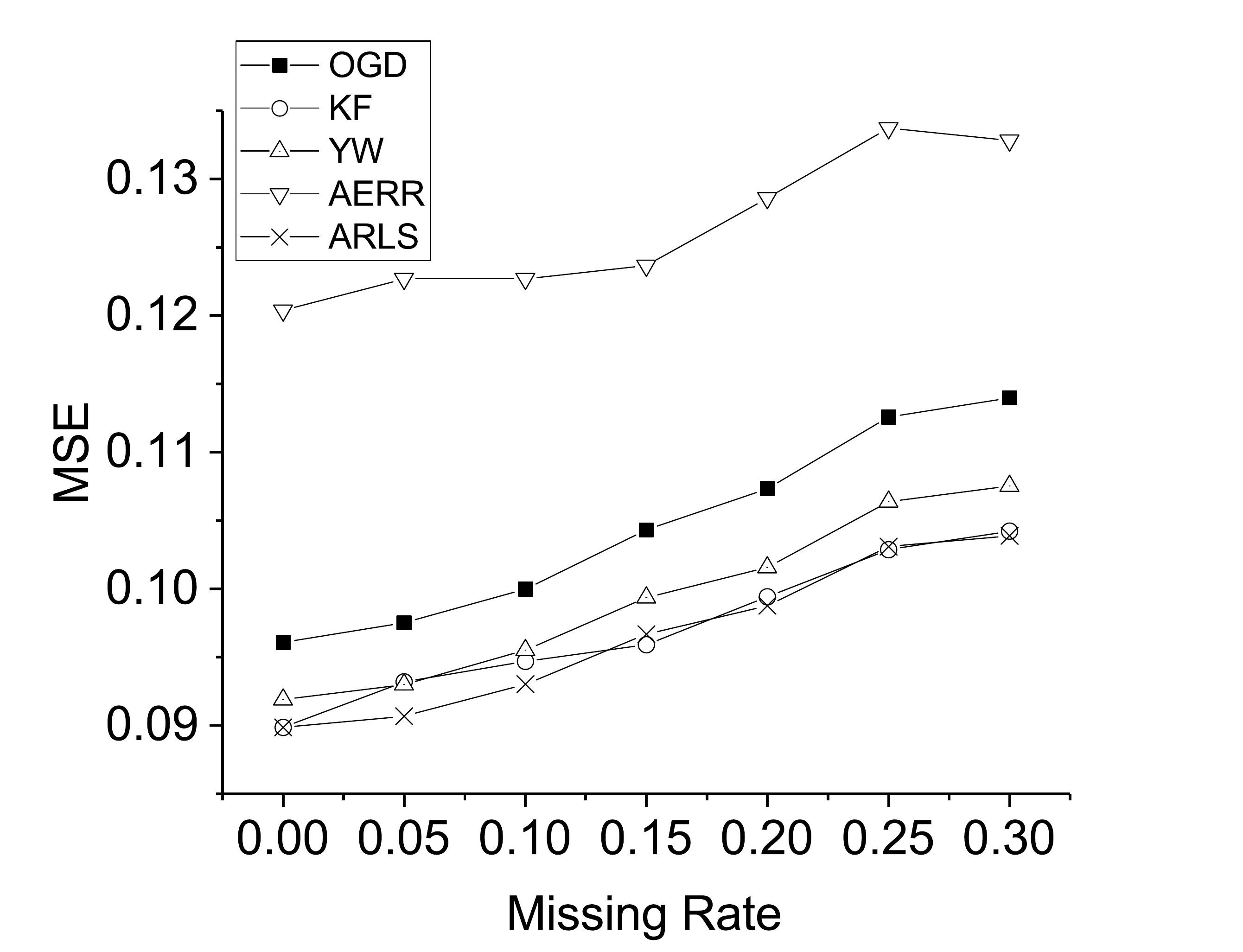}}
	\subfigure[Coefficients No.4]{
		\label{fig:ext3:4}
		\includegraphics[width=0.48\textwidth]{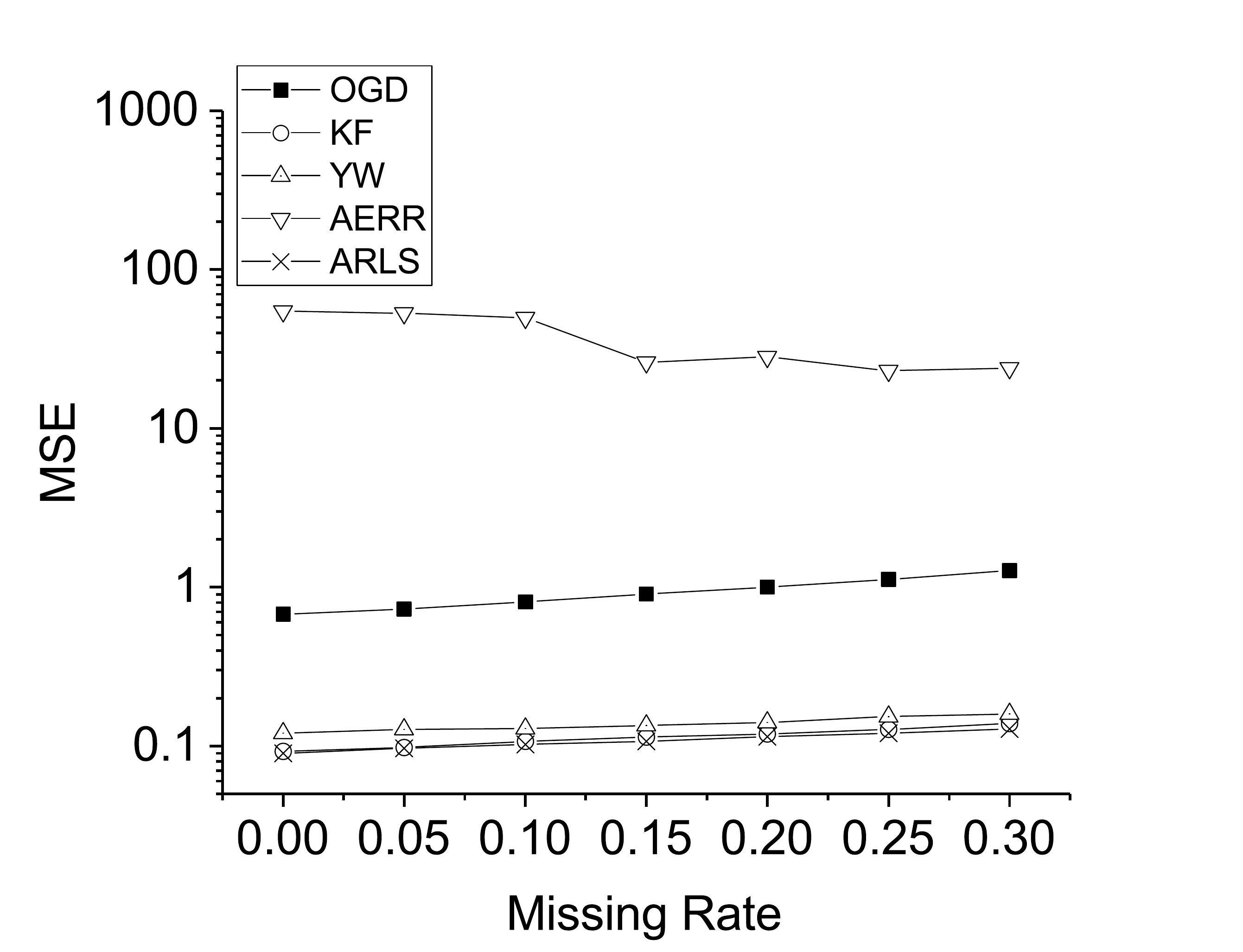}}
	\subfigure[Coefficients No.5]{
		\label{fig:ext3:5}
		\includegraphics[width=0.48\textwidth]{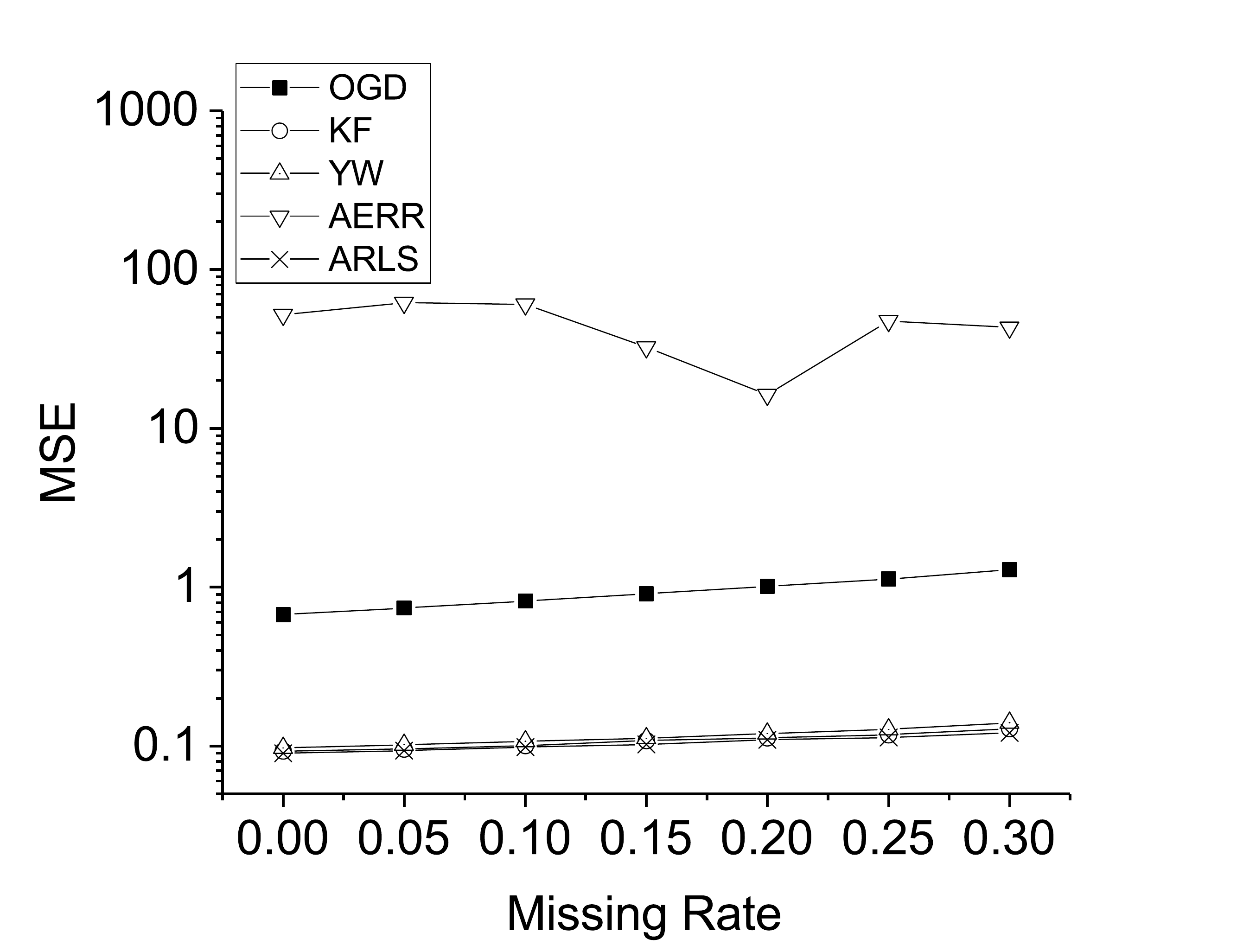}}
	\caption{MSE VS. AR Model Coefficients}
	\label{fig:ext3}
\end{figure}

In order to conduct comparisons on various types of synthetic data, we select five different AR model coefficients to generate time series data. The setting of coefficients is shown in Table \ref{AR_Coef_Setting} and the experimental results are shown in Figure \ref{fig:ext3}.

\begin{table}[H]
 \caption{Setting of AR Model Coefficients}
  \centering
  \begin{tabular}{cc}
    \toprule
     No.     & Coefficients  \\
    \midrule
    1 & [0.3, -0.4, 0.4, -0.5, 0.6]   \\
    2 & [-0.4, 0.4, -0.5, 0.3, 0.6]  \\
    3 & [0.3, 0.0, 0.5, 0.3, -0.2]  \\
    4 & [0.1, 0.7, 0.7, 0.0, -0.5] \\
    5 & [0.1, 0.7, 0.7, 0.0, -0.5] \\
    \bottomrule
  \end{tabular}
  \label{AR_Coef_Setting}
\end{table}

Comparing the five experimental results, it is observed that the prediction accuracies are relatively close to each other when the amplitude of the observation sequence is small (Figure \ref{fig:ext3:1} and \ref{fig:ext3:3}). However, only the KF, YW, and ARLS keep a low MSE when the sequence variation is large (which implies a nonstationarity of the sequence). Consistent with our analysis in the last experiment, the prediction errors of AERR are significantly more conspicuous when the sequence changes more sharply, which can be observed by comparing Figure \ref{fig:ext3:2} and \ref{fig:ext3:4}. For the OGD method, we tried various learning rates to perform a gradient descent. We found that when the sequence change over time is large, it requires a lower learning rate to ensure that it does not diverge. This has made OGD require more observations to allow the weight vector to converge. The results in Figure \ref{fig:ext3:2}, \ref{fig:ext3:4} and \ref{fig:ext3:5} show that the OGD method can adapt to a time series with a large variation, but converges more slowly than the KF and YW methods because the low learning rate limits the speed of optimization.

\subsection{Impact of Prediction Model Order}
\label{sec:predictmodelorder}
\begin{figure}[htbp]
	\centering
	\subfigure[$p_{fit}=1$]{
		\label{fig:ext4:1}
		\includegraphics[width=0.48\textwidth]{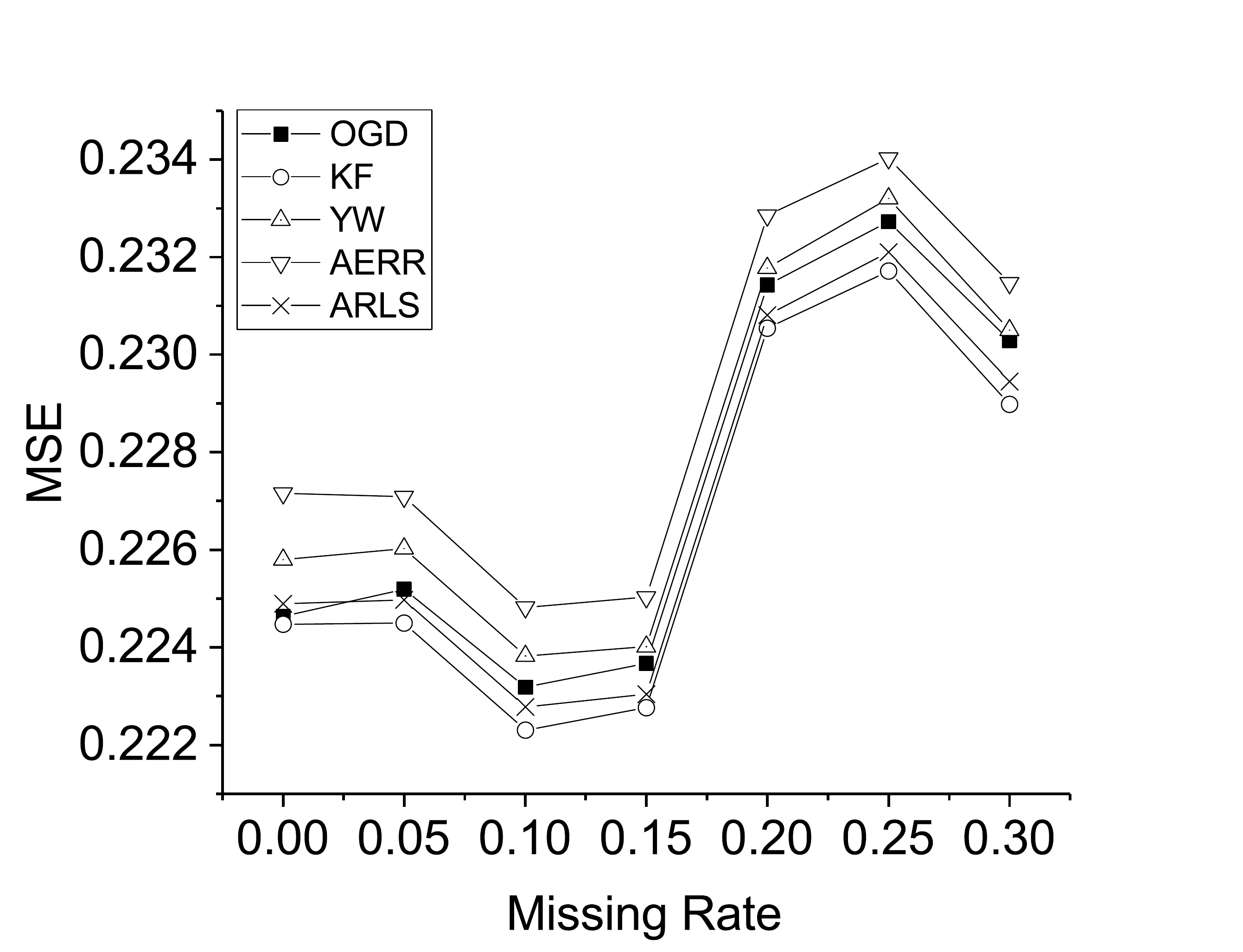}}
	\subfigure[$p_{fit}=5$]{
		\label{fig:ext4:5}
		\includegraphics[width=0.48\textwidth]{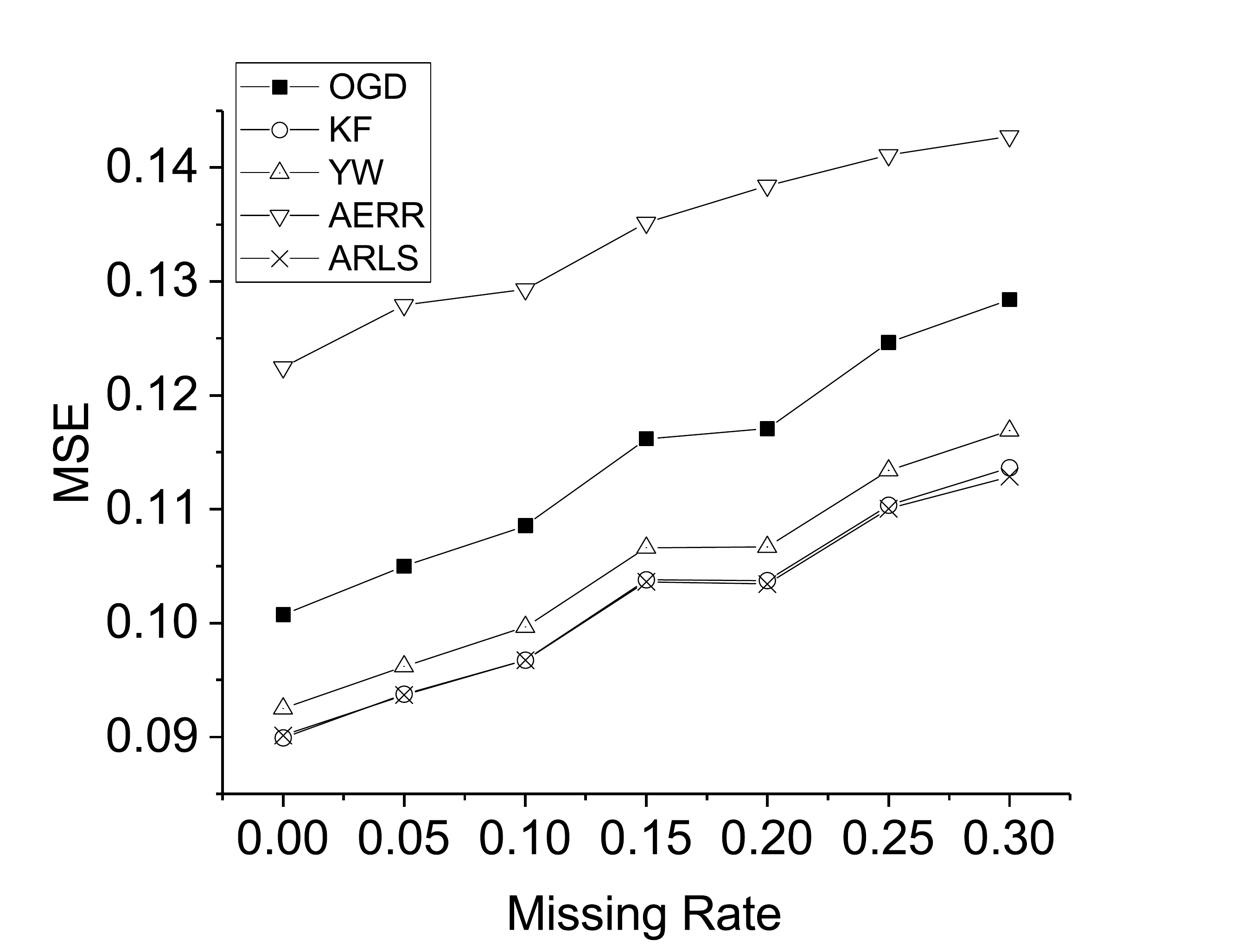}}
	\subfigure[$p_{fit}=20$]{
		\label{fig:ext4:20}
		\includegraphics[width=0.48\textwidth]{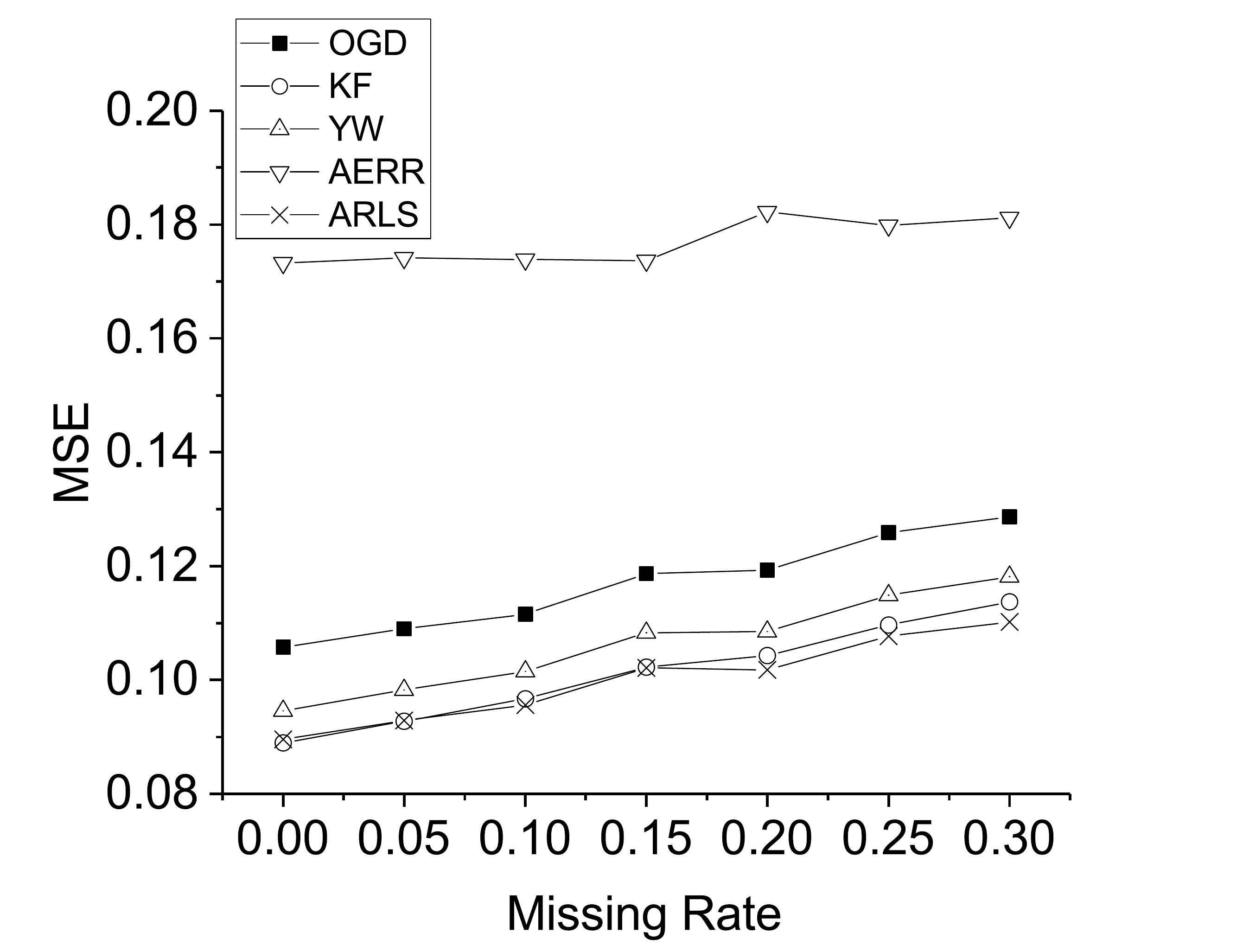}}
	\subfigure[$miss=0.05$]{
		\label{fig:ext4:0.05}
		\includegraphics[width=0.48\textwidth]{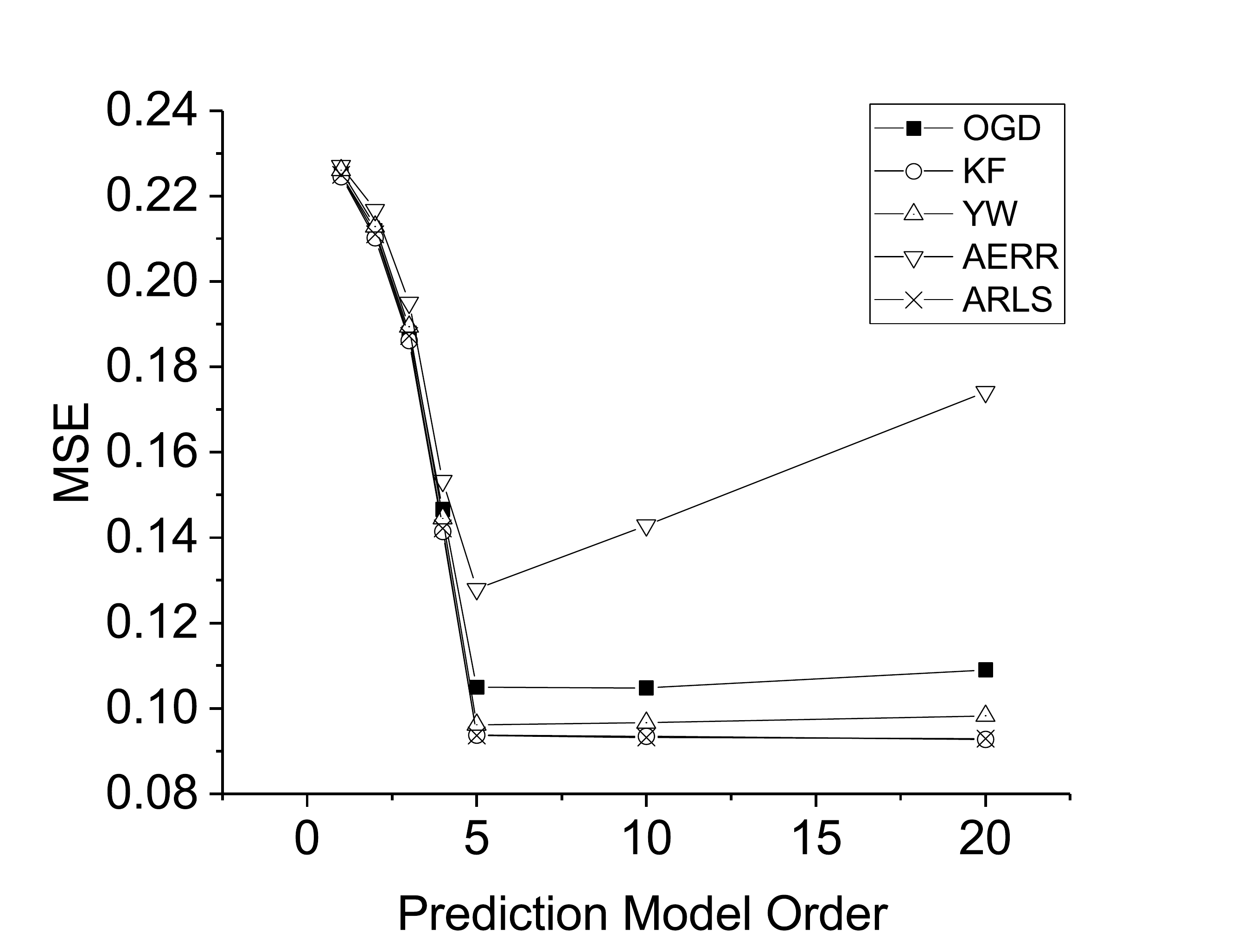}}
	\subfigure[$miss=0.3$]{
		\label{fig:ext4:0.3}
		\includegraphics[width=0.48\textwidth]{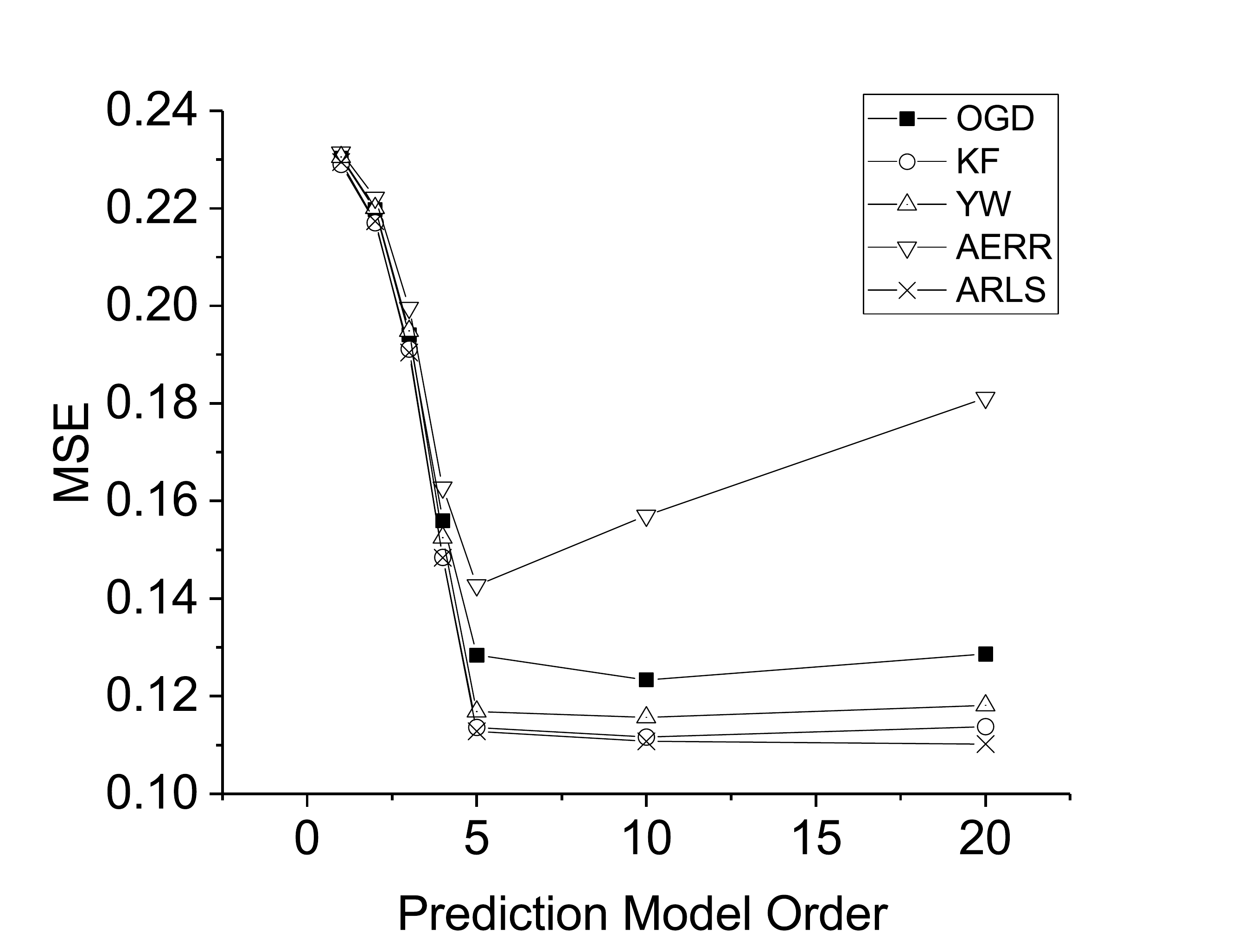}}	
	\caption{MSE VS. Prediction Model Order}
	\label{fig:ext4}
\end{figure}

In this experiment, we study the impact of prediction model order. which is used to determine the number of past observations used to predict the next observation. We set 7 values, \[p_{fit}=[1,2,3,4,5,10,20]\] For each $p_{fit}$ and each missing rate, we generate 20 time series for testing, and all methods run on these data for 7 times with 7 different prediction model orders. The experimental results are shown in Figure \ref{fig:ext4}.

From the experimental results, we observe that when the prediction model order $p_{fit}$ is lower than the AR model order $p_{gen}$ which represents the number of AR coefficients, the MSEs of different methods have different degrees of decrease with the increase of $p_{fit}$, and their changing trend with respect to the missing rate is closer to a linear growth.

When $p_{fit}$ exceeds $p_{gen}$, the prediction performance of OGD, YW, KF and ARLS does not change significantly, but the prediction errors of AERR increase. With regard to this result, when the order of the prediction model increases, in order to ensure the weight vector converges, all these methods require more observations for the training.

For AERR, the increase in the demand for data exceeds the amount of available data, so that the predictions of AERR are based on the estimation of AR model coefficients that have not converged, resulting in an increase in MSE. Additionally, note that the prediction performance of other methods is not improved by increasing the order of the model, indicating that higher order of the prediction model would not lead to better results.

The results of this experiment suggest that time series prediction based on sampling and gradient descent is not suitable for high-order model training and should not be applied to applications where the amount of data available is small.

\subsection{Performance on Real Data}
\label{sec:realdata}
To comprehensively compare these methods, we conduct experiments on real data as well. Compared to synthetic data, the real data is not generated by a simple static AR model. Besides, the feature of its underlying noise is unknown.
Observing the experimental results, the difference between performance of different methods is obviously larger than what was shown in the experimental results on synthetic data.

To make it more clear about the experimental data, we show brief information in Table \ref{RealData_Info}. There are four sets of real data. The data of \emph{Stock} are collected from \emph{TuShare} \footnote{\url{http://tushare.org/}} and others are all from \emph{UCR Archive} \footnote{\url{http://www.cs.ucr.edu/~eamonn/time_series_data/}}. Missing values in these real time series are generated by setting the observation value to empty according to a certain missing rate for each time point.

\begin{table}[htbp]
	\centering
	\caption{Real Data Profile}
	\label{RealData_Info}
	\begin{tabular}{ccc}
		\toprule
		Category & Number of Groups & Length \\
		\midrule
		Stock & 40 & 1000 \\
		Coffee & 2 & 286 \\
		Inline Skate & 6 & 1882 \\
	    Synthetic Control & 6 & 60 \\
	    \bottomrule
	\end{tabular}
\end{table}

\subsubsection{Stock}
\label{sec:stock}
\begin{figure}[htbp]
	\centering
	\subfigure[MSE of all methods]{
		\label{fig:std2:1}
		\includegraphics[width=0.48\textwidth]{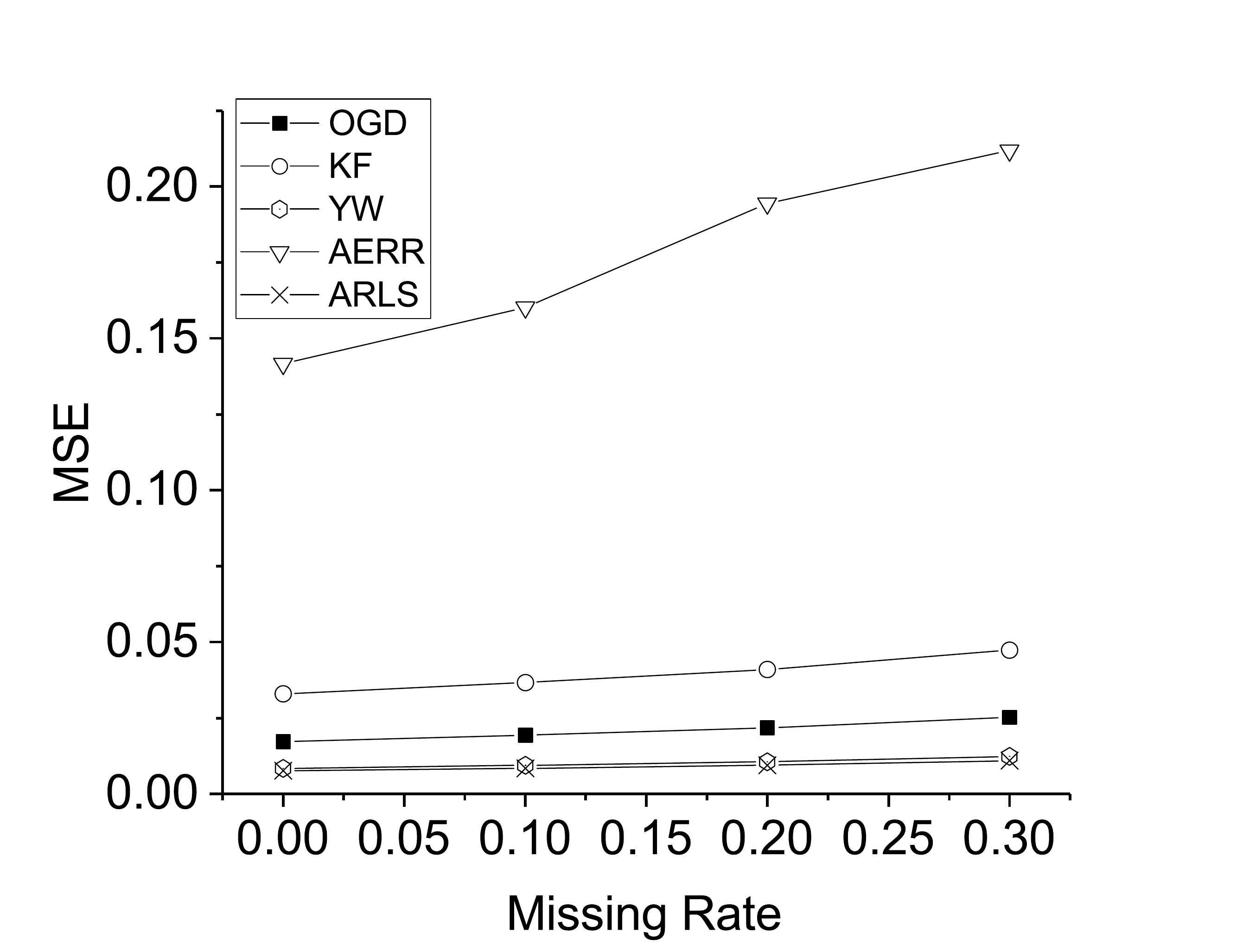}}
	\subfigure[MSE excluding AERR]{
		\label{fig:std2:2}
		\includegraphics[width=0.48\textwidth]{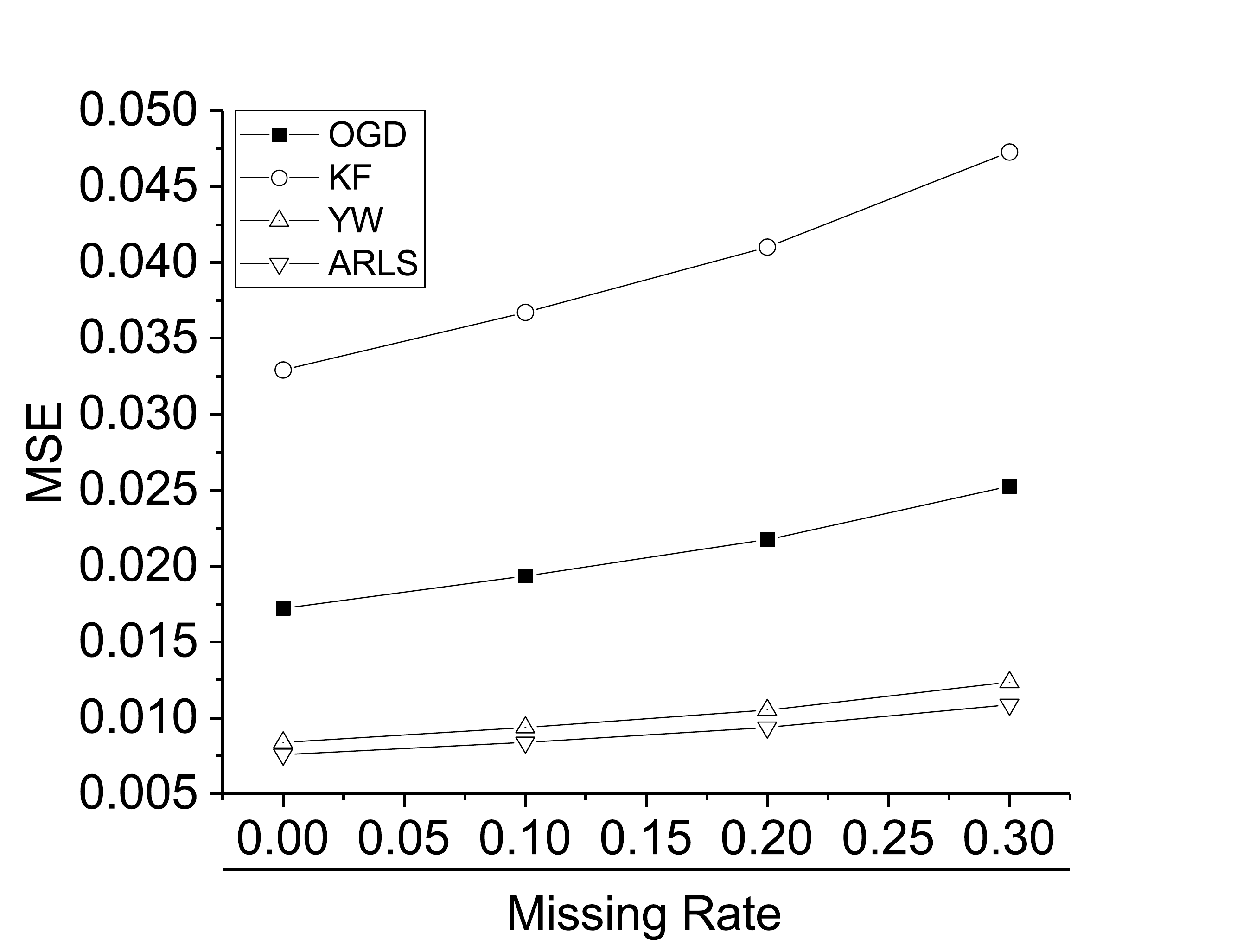}}
	\caption{ Experimental Results on \emph{Stock}}
	\label{fig:std2}
\end{figure}

In this experiment, we test the performance of these methods on the time series of stock prices. In terms of data source, we first obtain the raw data of 5 years stock price from \emph{TuShare}. Then we chose 40 stocks and extracted only the close price of every trading day to make up 40 time series. For the convenience of computing, we preprocessed them by only keeping the first 1000 observations in each time series and conducting z-score to normalize the data. The experimental results are shown in the Figure \ref{fig:std2}.

As shown in the Figure \ref{fig:std2:1}, the difference among all methods become more obvious. AERR, in this case, does not achieve a satisfactory result. When the missing rate is 0.3, the MSE of AERR even gets up to over 10 times that of ARLS.

To analyze this phenomenon, we need to briefly review the core idea of AERR. The AERR method has double sampling, a uniform sampling and a sampling on the basis of the latest weights (AR coefficients). The original AERR is under any-time setting. we adapt it to the online version according to following steps. Firstly, we perform the double sampling process without repairing missing observations. In this way, we update the weights without being affected by possible previous bad predictions. Secondly, we use the new weights and the current observation vector (the latest $p$ observations or imputed values) to make more stable predictions. Considering the time cost, the sampling time $k$ is not always good to be large. By testing on several numbers, we set a proper sampling times $k=10$ to achieve both accuracy and efficiency.

However, from the experiment results, AERR performs not well especially under the real data setting. One reason is that for real data, the characters of data are hard to know. Thus, the parameters are difficult to determine. Since AERR requires more parameters, it is hardly to obtain proper parameters, which causes low accuracy. Another point is that previous predictions is, to some extent, a compressed form of historical information. The other online methods use both the latest weights and former predictions to make a new prediction, but AERR discards the latter, which makes AERR unable to benefit from those information.

In order to make it more clear for the rest methods, we plot MSEs of these methods in Figure \ref{fig:std2:2}. From the results, KF performs not so ideal. Its MSEs stand as the highest over that of other methods, and they have the most dramatic growing trend. However, this is because of its intrinsic assumption of a Gaussian white noise. The dependency on noise distribution limits its performance in real data situation.

As for OGD, it shows a more stable outcome. Despite the fact that its MSEs still approximately double that of YW and ARLS, the overall stability and fast speed are strong support for choosing this method in most of the online prediction situations.

To our surprise, YW outperforms all other online methods. From the results, we observed that YW tends to make the predicted value gradually approach the actual observations while OGD takes a more direct way by trying to stick to the current observation as soon as possible. This is the reason for the different performance between these two methods. Since YW needs to acquire the autocorrelation coefficients with the accumulation of time, the learning process naturally appears to be slow. However, when the data set gets to be large enough, it can infer the future observation with more accuracy. In contrary, OGD does not explicitly store historical information, and its target is to optimize the AR model based on the current observation. This is more like a greedy strategy, so the updating of weight is more easily affected by sudden changes in time series. As a result, the average prediction error of OGD gets larger.

\subsubsection{Coffee}
\label{sec:coffee}
\begin{figure}[htbp]
	\centering
	\subfigure[Group No.1]{
		\label{fig:ext_coffee:0}
		\includegraphics[width=0.48\textwidth]{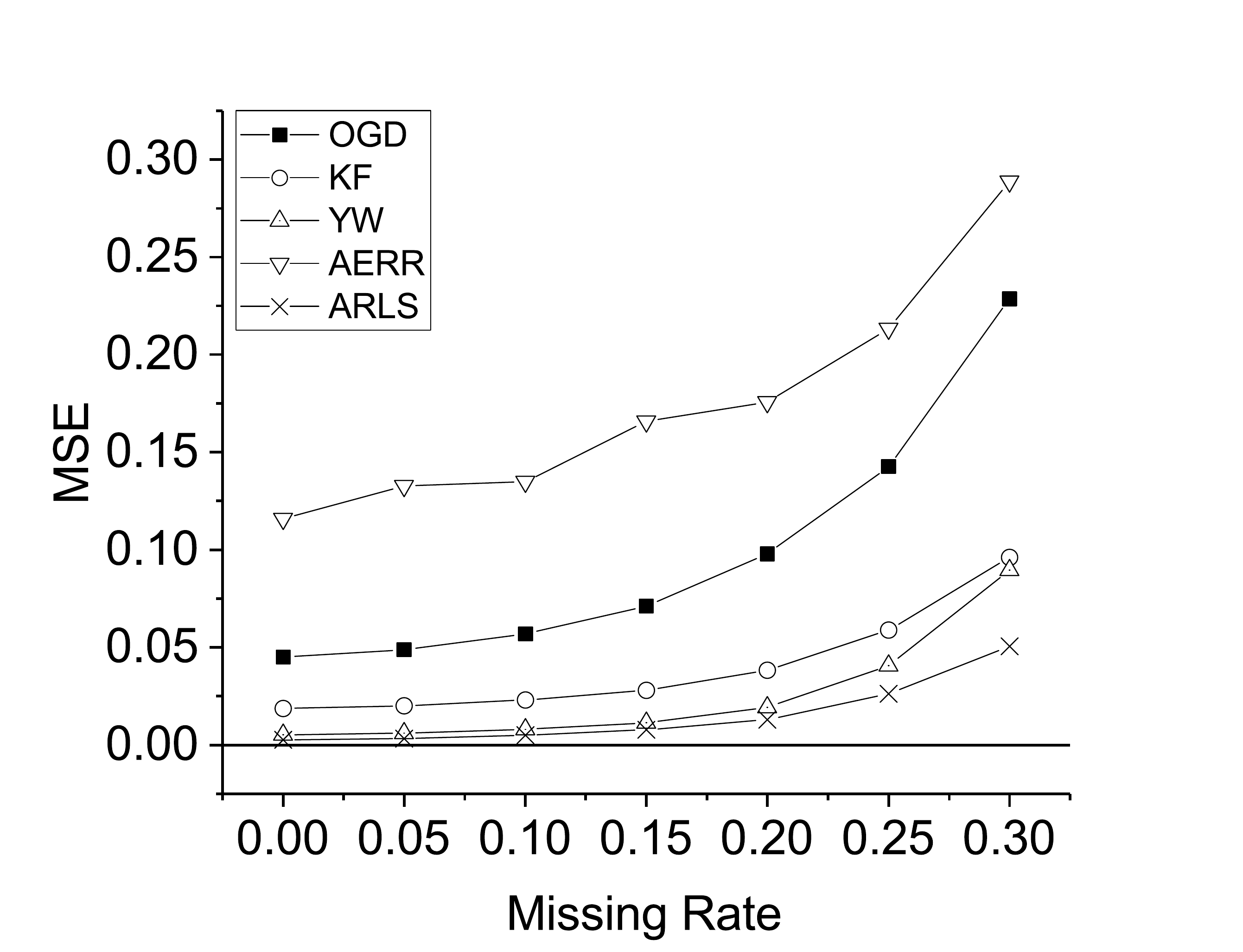}}
	\subfigure[Group No.2]{
		\label{fig:ext_coffee:1}
		\includegraphics[width=0.48\textwidth]{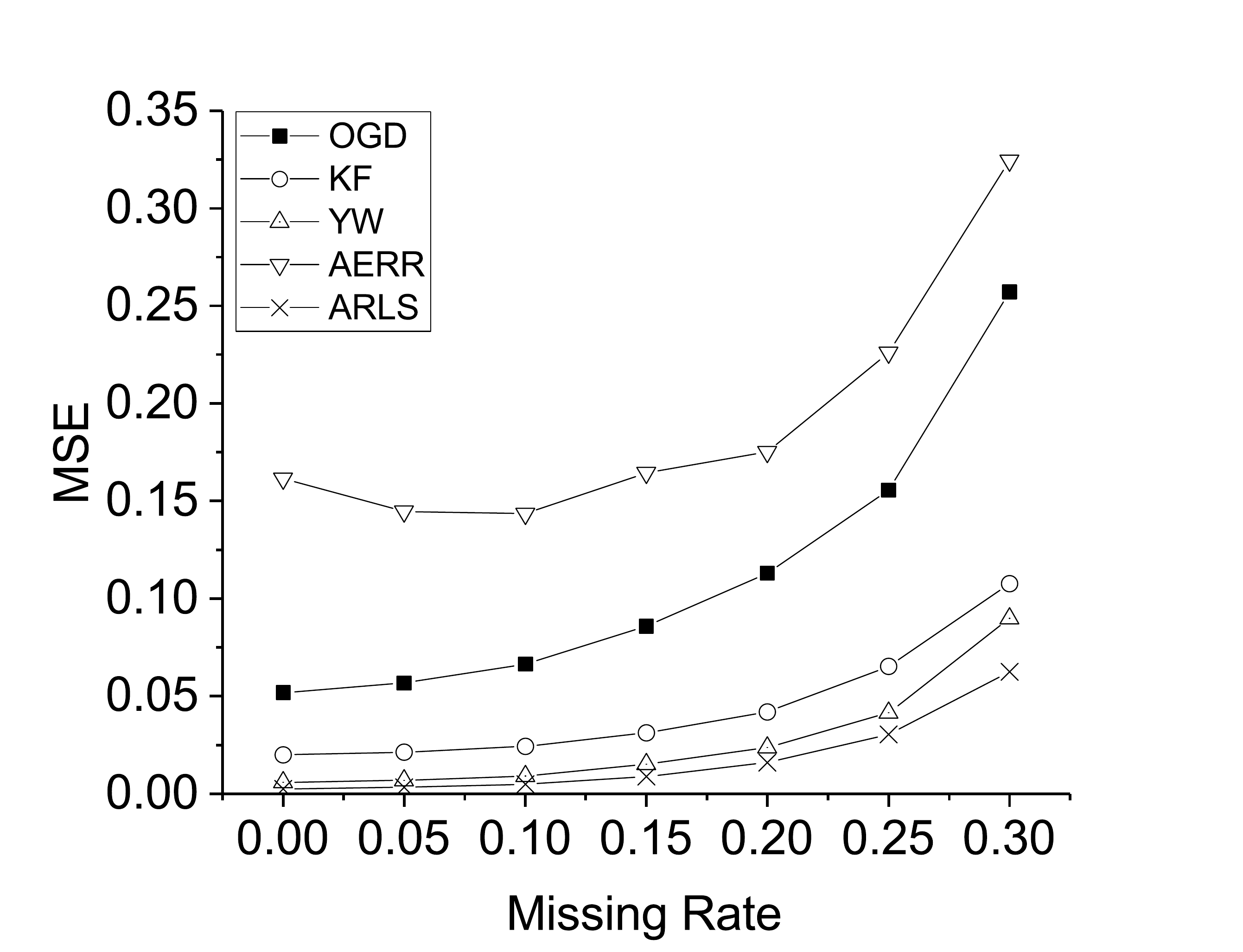}}
	\caption{Experimental Results on \emph{Coffee}}
	\label{fig:ext_coffee}
\end{figure}

The length of the time series in \emph{Coffee} is 286. This class of data contains two groups of time series with nuances. The observation values roughly revolve around zero, and the track of their changing over time is similar to the contour of a mountain. The experimental results are shown in Figure \ref{fig:ext_coffee}.

According to these figures, KF, YW and ARLS consistently maintain the leading edge under all the missing rates, and their MSEs' ascending trend is more gentle. In contrast, AERR has a large MSE at the zero missing rate, and its MSE grows faster as the missing rate increases. It is noteworthy that OGD performs well at low missing rates, but with the increase of the data loss, its MSE rapidly approaches the worst MSE of AERR. This observation indicates that OGD is more sensitive to the absence of data in this type. This is due to the fact that the stochastic gradient descent method is susceptible to the randomness of the data, and the use of the imputation strategy to deal with the missing values leads to the accumulation of prediction errors. The effect of this accumulation is particularly pronounced when the missing rate is large, and the time series changes rapidly and severely.

\subsubsection{Inline Skate}
\label{sec:inlineskate}
\begin{figure}[htbp]
	\centering
	\subfigure[Group No.1]{
		\label{fig:ext_inlineskate:1}
		\includegraphics[width=0.48\textwidth]{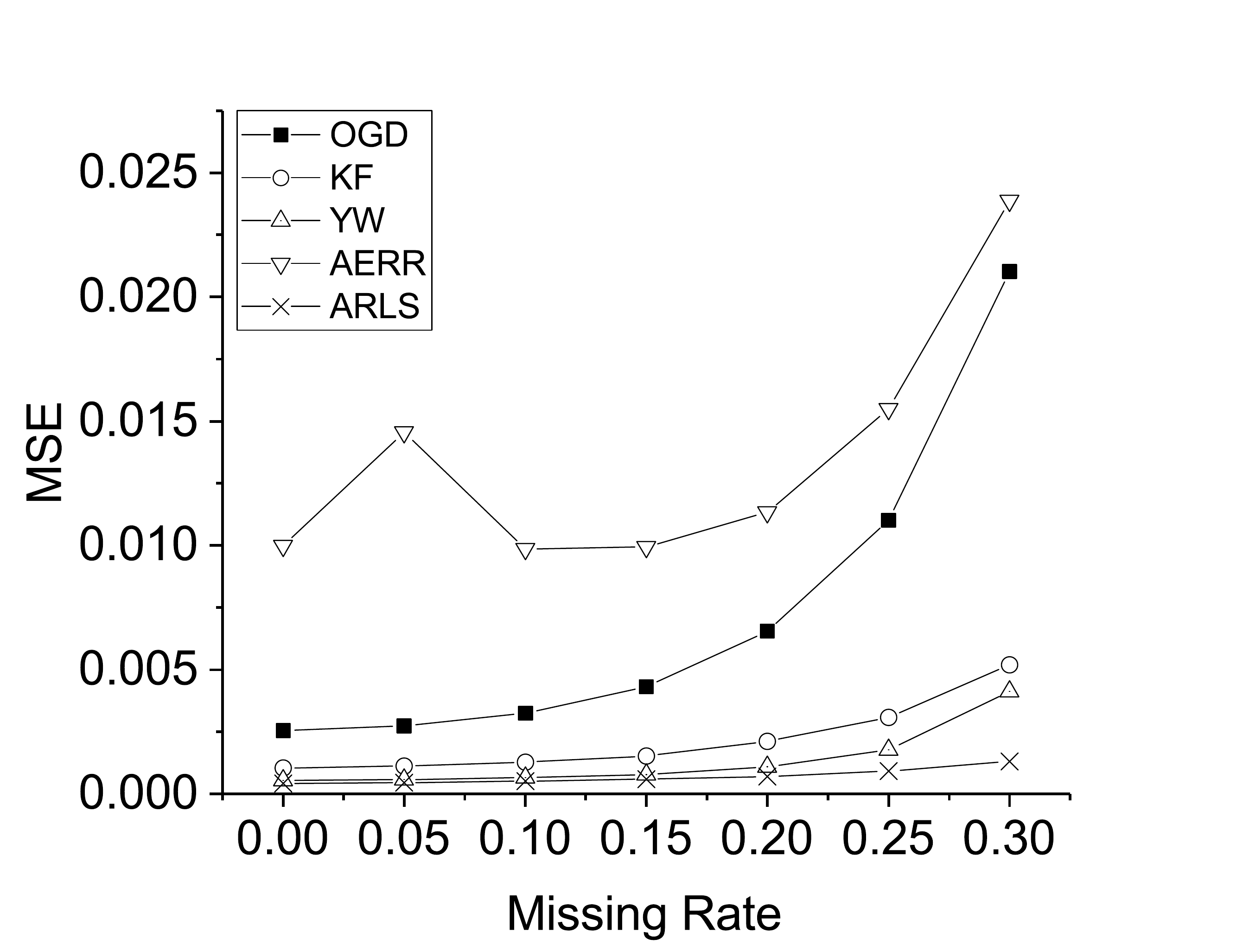}}
	\subfigure[Group No.2]{
		\label{fig:ext_inlineskate:2}
		\includegraphics[width=0.48\textwidth]{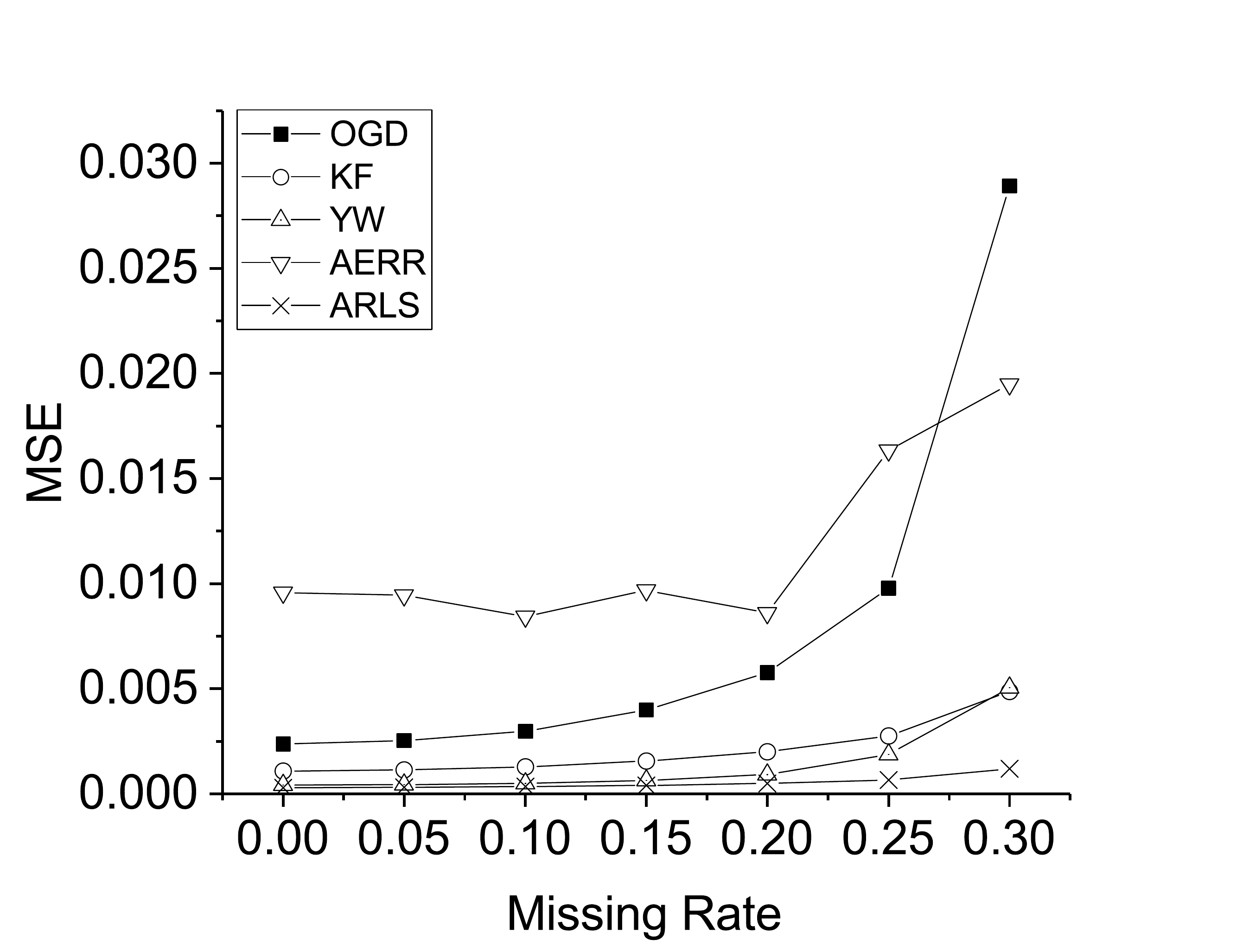}}
	\subfigure[Group No.3]{
		\label{fig:ext_inlineskate:3}
		\includegraphics[width=0.48\textwidth]{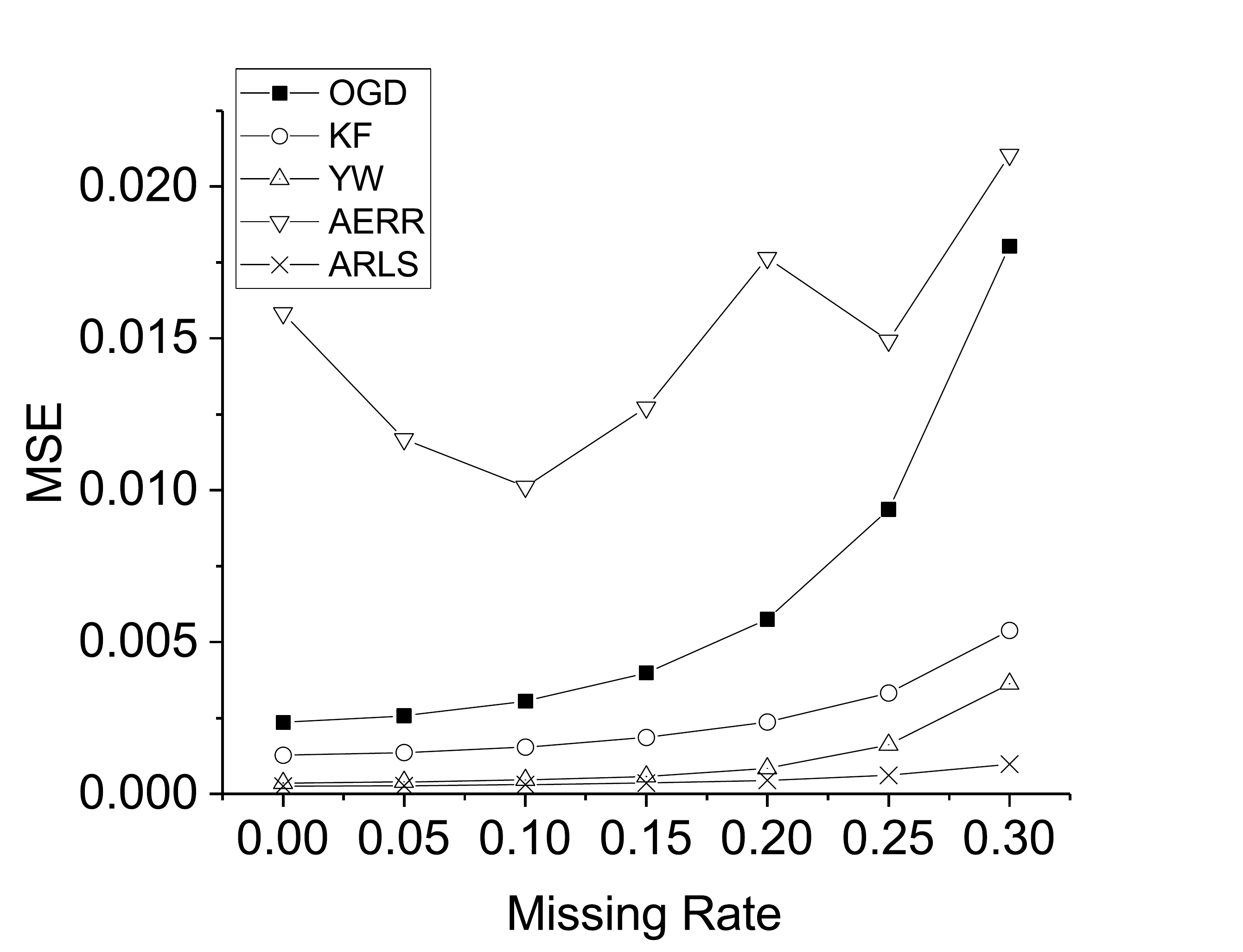}}
	\subfigure[Group No.4]{
		\label{fig:ext_inlineskate:4}
		\includegraphics[width=0.48\textwidth]{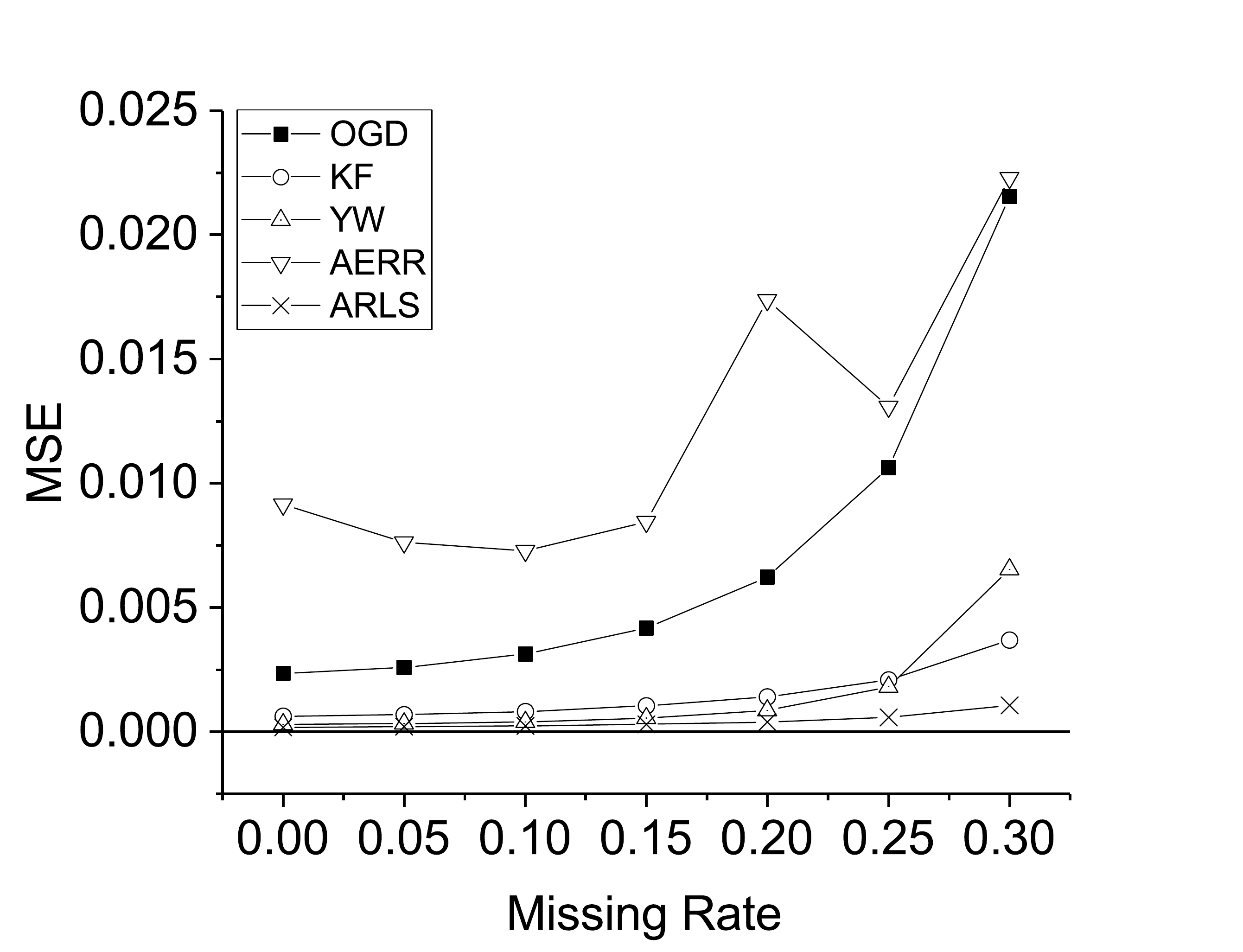}}
	\subfigure[Group No.5]{
		\label{fig:ext_inlineskate:5}
		\includegraphics[width=0.48\textwidth]{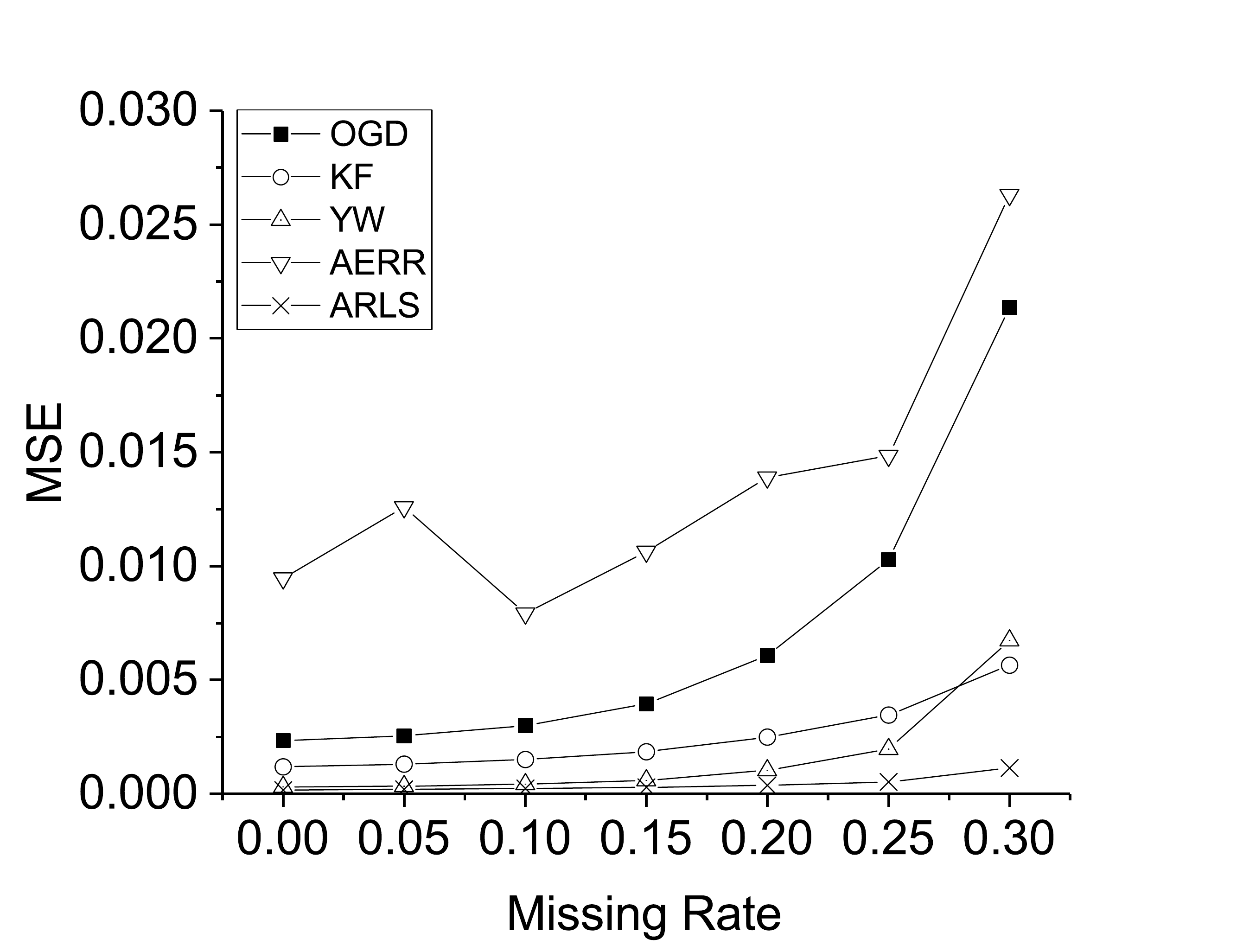}}
	\subfigure[Group No.6]{
		\label{fig:ext_inlineskate:6}
		\includegraphics[width=0.48\textwidth]{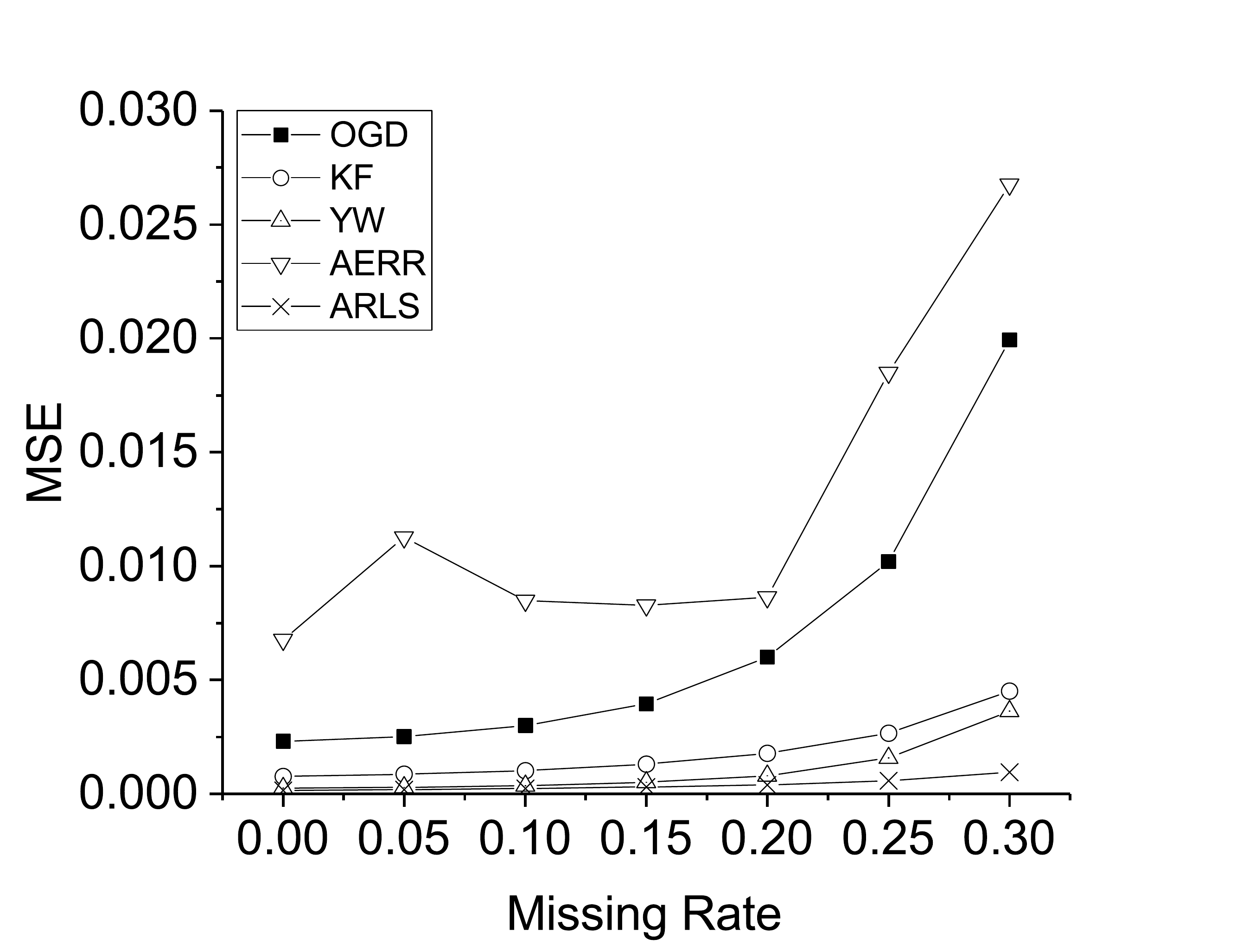}}
	\caption{Experimental Results on \emph{Inline Skate}}
	\label{fig:ext_inlineskate}
\end{figure}

Similar experimental results in Figure \ref{fig:ext_inlineskate} appear on \emph{Inline Skate} which contains 6 groups  of time series with the length of 1882. The time series is also mountain-like, but the ups and downs of observation values are more drastic, the time span is longer and the sequence overall is smoother. From the experimental results, the MSE of OGD exceeds that of AERR at the maximum missing rate, with the largest prediction deviation, which is consistent with our previous speculation.

In addition, experimental results of current two types of data also show that the results of ARLS always keeps the lowest prediction error, while YW and KF are still ideal. The MSE of KF is slightly higher than that of YW, but the latter increases faster as the missing rate increases. In the experimental results of the data group 4 and 5 in Figure \ref{fig:ext_inlineskate:4} and \ref{fig:ext_inlineskate:5}, its MSE significantly exceeds that of the KF method at the missing rate of 0.3. This result suggests that the KF method is more tolerant of data missing than the YW method.

\subsubsection{Synthetic Control}
\label{sec:syncon}
\begin{figure}[htbp]
	\centering
	\subfigure[Group No.1]{
		\label{fig:ext_syncon:1}
		\includegraphics[width=0.48\textwidth]{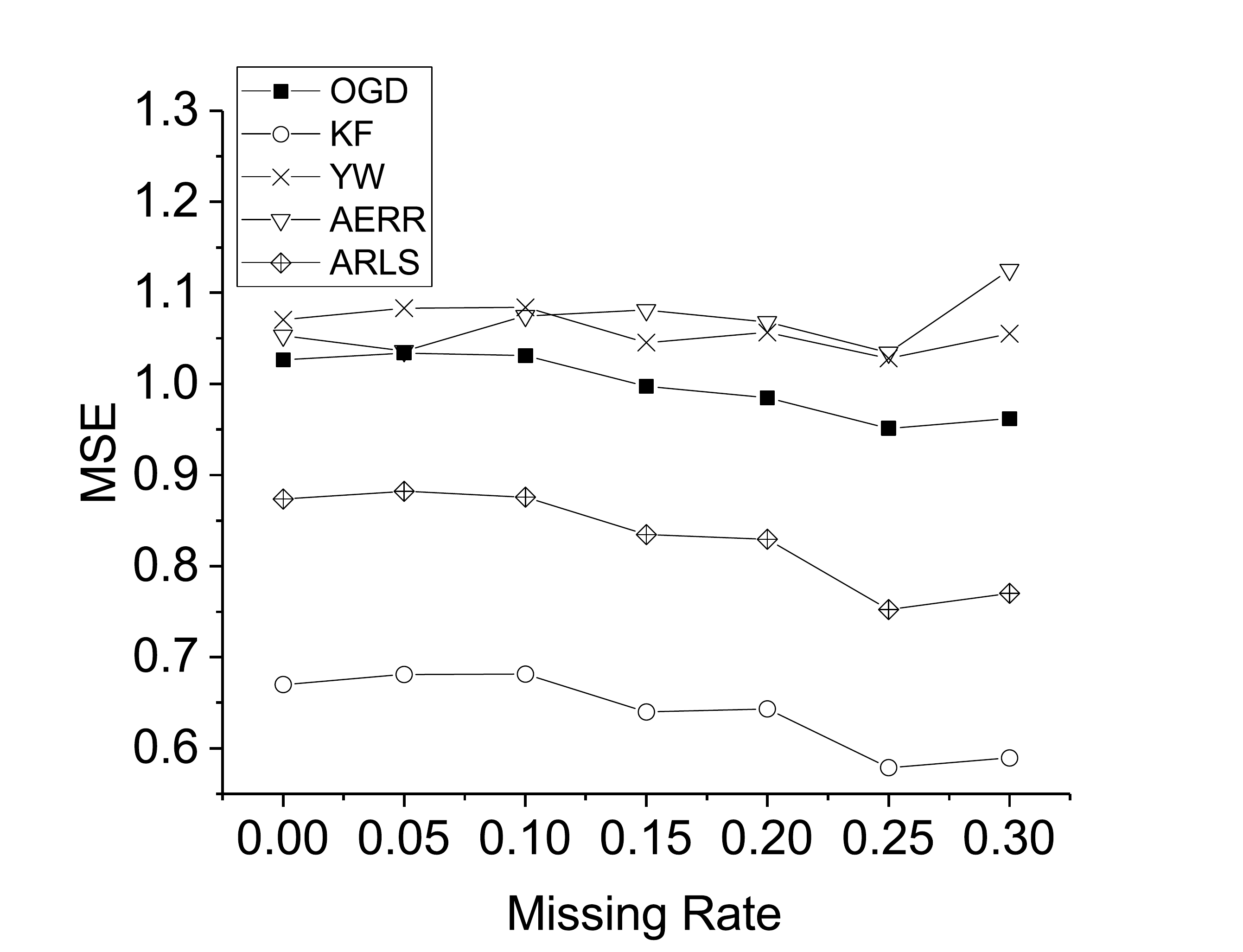}}
	\subfigure[Group No.2]{
		\label{fig:ext_syncon:2}
		\includegraphics[width=0.48\textwidth]{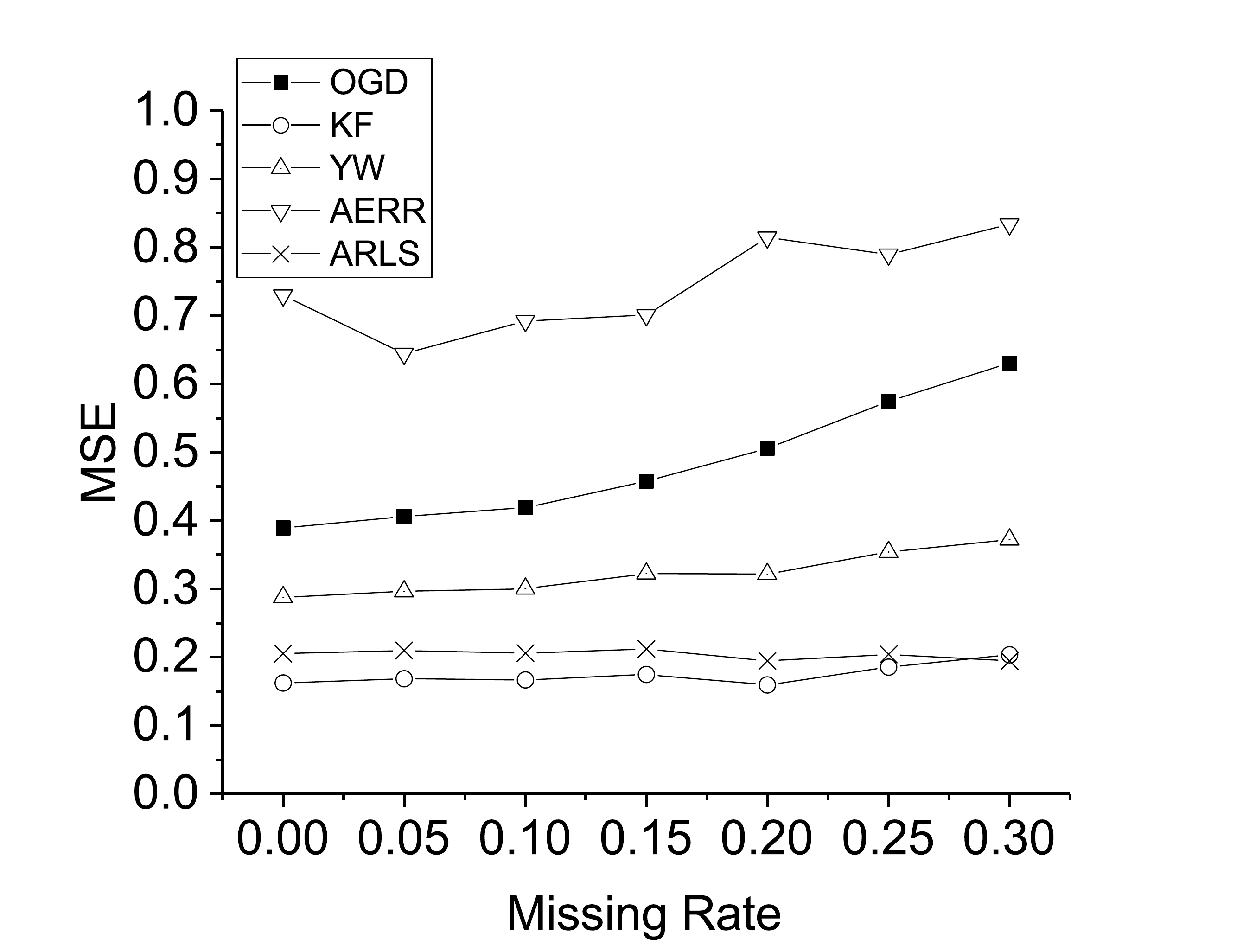}}
	\subfigure[Group No.3]{
		\label{fig:ext_syncon:3}
		\includegraphics[width=0.48\textwidth]{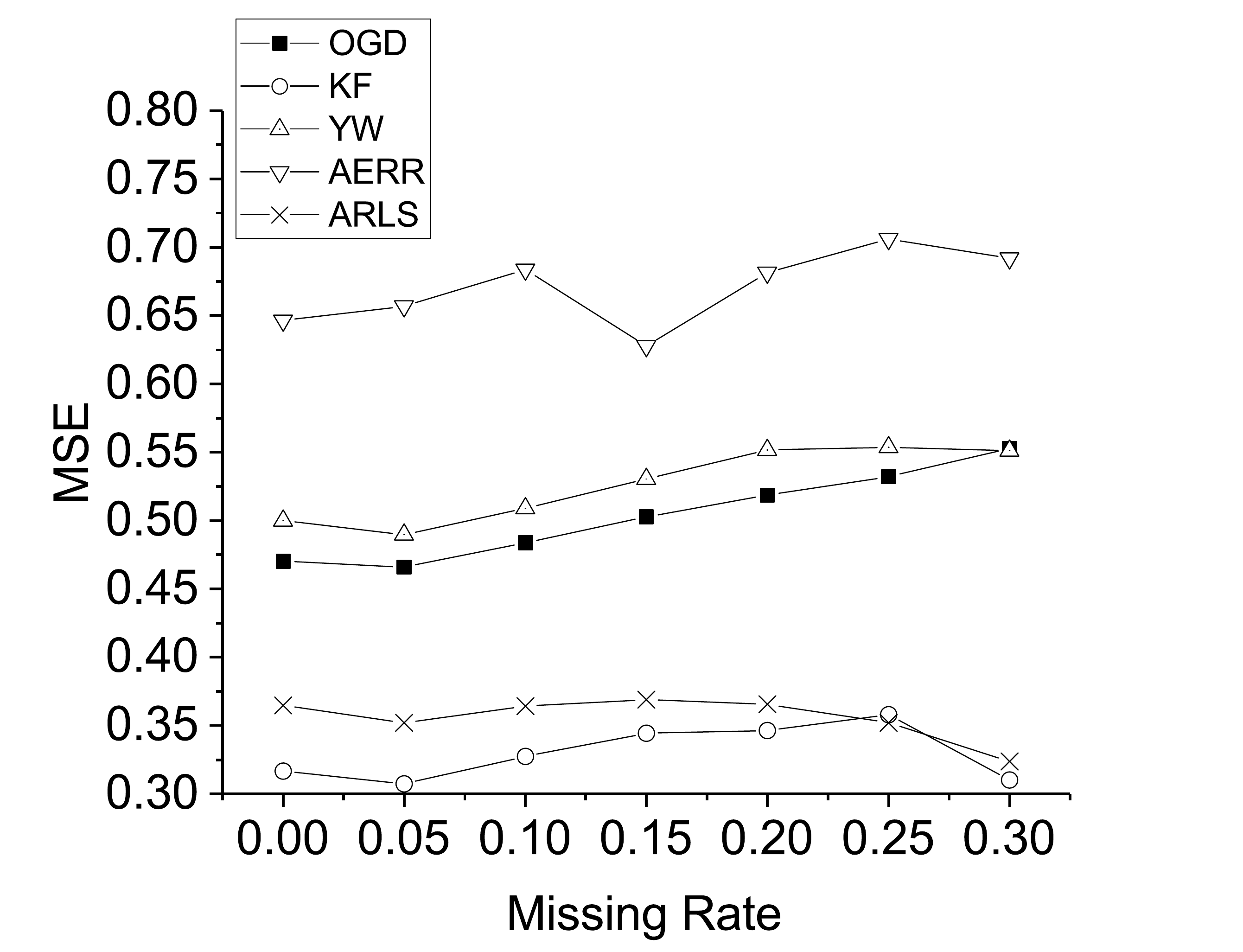}}
	\subfigure[Group No.4]{
		\label{fig:ext_syncon:4}
		\includegraphics[width=0.48\textwidth]{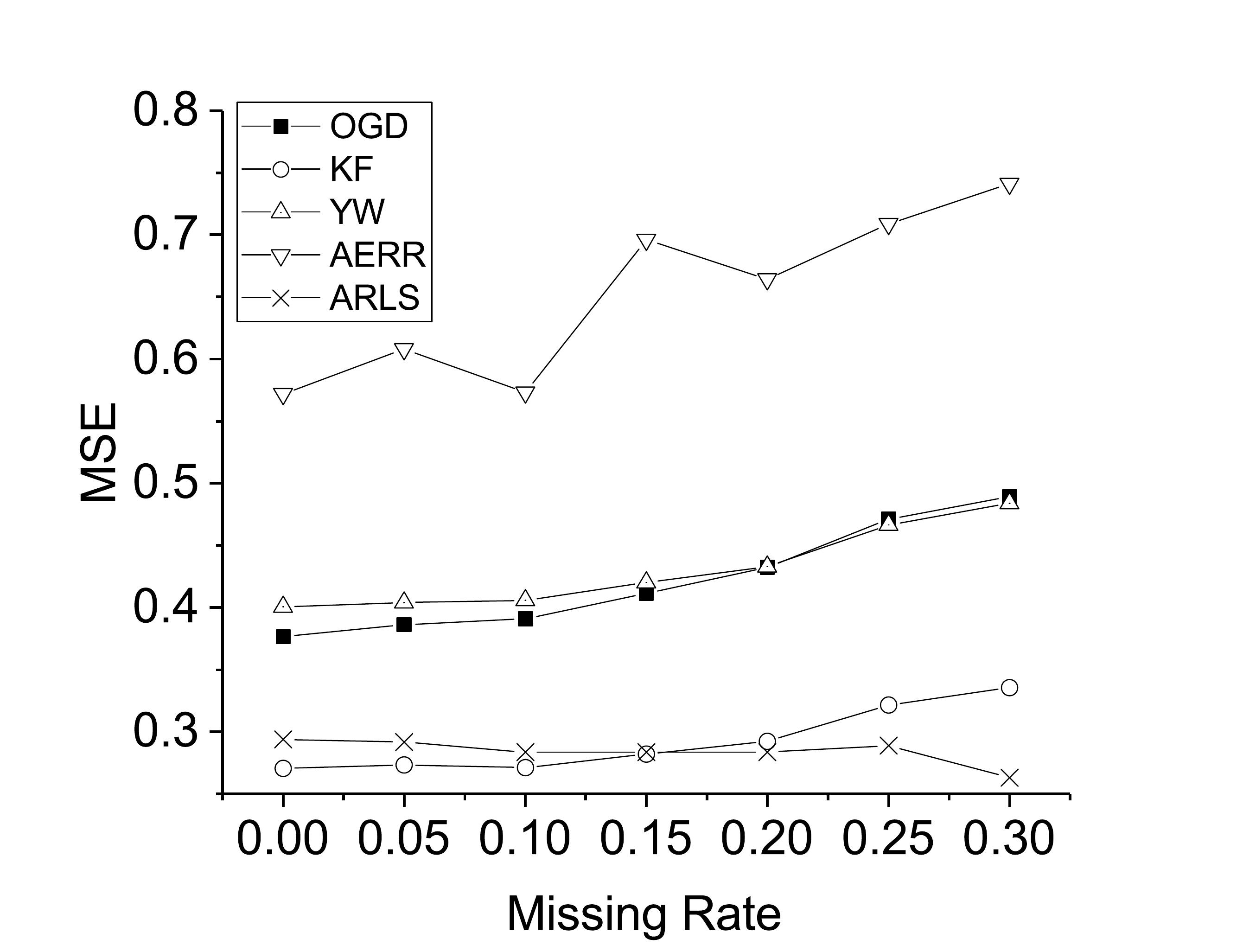}}
	\subfigure[Group No.5]{
		\label{fig:ext_syncon:5}
		\includegraphics[width=0.48\textwidth]{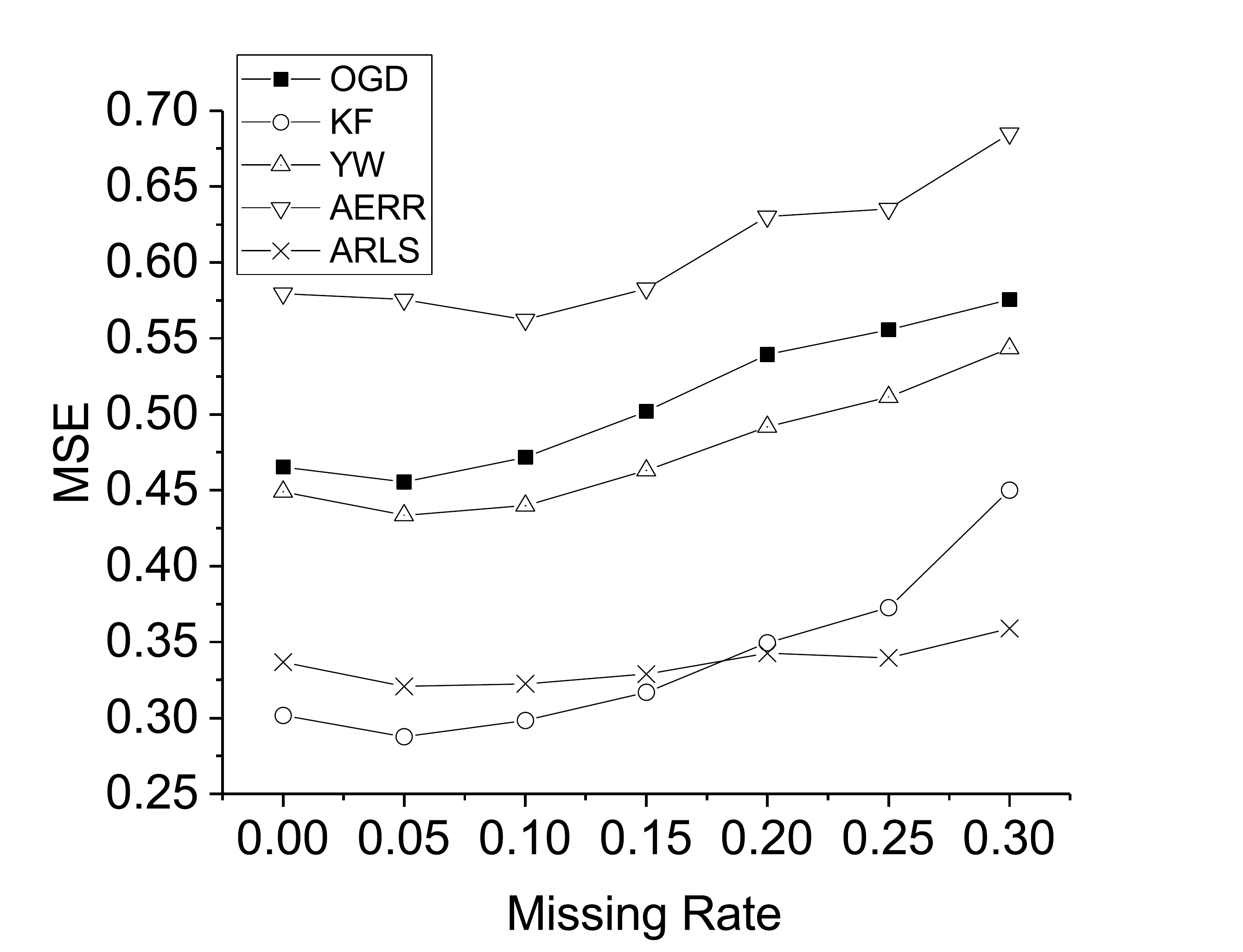}}
	\subfigure[Group No.6]{
		\label{fig:ext_syncon:6}
		\includegraphics[width=0.48\textwidth]{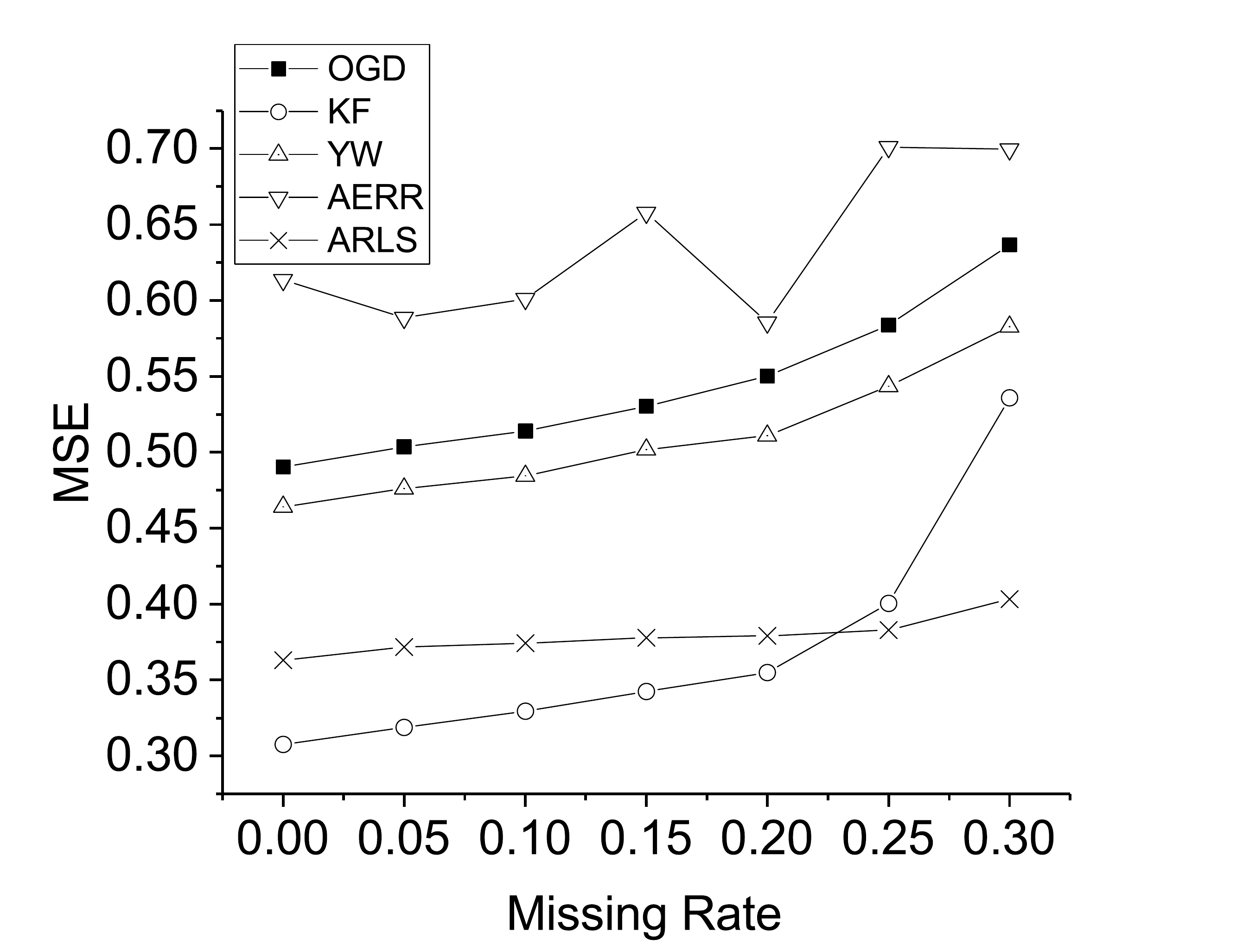}}
	\caption{Experimental Results on \emph{Synthetic Control}}
	\label{fig:ext_syncon}
\end{figure}

For the experimental results in Figure \ref{fig:ext_syncon} derived from \emph{SyntheticControl}
which contains 6 groups of time series with a length of 60, we obtain some different observations. First, different from the first two types of time series, the length of this type of time series is shorter, and there are obviously different characteristics between the groups of time series. For example, the first and second groups of data are sequences that are similar to white noise. However the changes of the observation values are more intensive in the first one, while the seasonality in the second one is remarkable. There are obvious rising or falling trends in the remaining groups of data. Secondly, it is observed from the experimental results of the first group of data that the OGD, AERR and YW methods share similar but not ideal prediction performance, while the KF method performs well. The poor performance of the YW method suggests that it is not suitable for such type of time series. This is because the core of the YW method is based on the estimation of the autocorrelation coefficient of the time series, and the autocorrelation of data within the white-noise-like time series is weak, which is not conducive to the YW method to obtain information from data. Moreover, compared with the experimental results of the second group of data, YW achieve good performance when there is an obvious periodicity in the time series, which coincides with our expectations.

In the remaining experiments, the four groups of time series are with a rising or falling trend or sudden changes. In spite of this, KF still exhibits a strong adaptive ability, and the rising or falling trend of the sequence does not have a significant effect on it. It has achieved a very good prediction results in all four groups of data. In contrast, YW behaves not ideal, and sometimes its MSEs get even higher than that of OGD. This is because its predictive ability depends on the assumption of stationarity that the autoregressive model makes on the time series. In addition, OGD and AERR keep a relatively high MSE, which is related to the slow convergence of the stochastic gradient descent algorithm on the data with noise.

\subsection{Method Comparison Summary}
From the experimental results, we summarize the following conclusions.

\begin{list}{\labelitemi}{\leftmargin=1em}\itemsep 0pt \parskip 0pt
\item The higher the missing rate, the greater the prediction error. Although the growing trend is closer to linear growth in case of synthetic data, it is generally non-linear in real data situation. This can be concluded from the experimental results on missing rates and the real data in Section \ref{sec:missingrate} and \ref{sec:realdata}.
\vspace{2mm}
\item The longer the time series, the smaller the difference between performance of different methods. It is observed from the experimental results on varying time series length, as is demonstrated in Section \ref{sec:tslength}.
\vspace{2mm}
\item AERR is more sensitive to the noise variance than other methods. When the variance is high, AERR tends to get bad performance, as shown in the experimental results of the impact of noise variance in Section \ref{sec:noisevariance}.
\vspace{2mm}
\item Both AERR and OGD have weak ability on tracking the rapidly changing time series, and AERR is especially bad for time series with large range. This can be concluded from the results of the experiment on AR model coefficients in Section \ref{sec:armodelcoeff}.
\vspace{2mm}
\item From the experimental results on the impact of prediction model order in Section \ref{sec:predictmodelorder}, the best order of prediction model is the actual order of the underlying AR model.
\vspace{2mm}	
\item The performance of OGD and AERR is generally poor, and they are more sensitive to the lack of data when the variation range of time series is large. This is concluded from the result of the experiments on \emph{Coffee} and \emph{Inline Skate} data sets in Section \ref{sec:coffee} and Section \ref{sec:inlineskate} respectively.
\vspace{2mm}	
\item KF and YW have generally better performance. However, YW method is not suitable for time series with no seasonality or with unstable trend, and KF has poorer performance in the case of the noise is not Gaussian, as the conclusion from the experimental results on \emph{Stock} and \emph{Synthetic Control} in Section \ref{sec:stock} and Section \ref{sec:syncon} respectively.

\end{list}

In summary, the poor performance of AERR may have implied the weakness of sampling-based method in online learning, and imputation is still a simple but reliable and effective way to cope with missing values in online prediction tasks.

\section{CONCLUSIONS}
In this paper, we compare the online prediction methods for the time series with missing values experimentally. Such problem is crucial for many applications since online prediction for time series has wide applications, and missing values are common in real time series.

Based on the assumption of the AR model on time series, we compared five different methods, including three imputation-based methods Yule-Walker Estimator (YW), Kalman Filter (KF), Online Gradient Descent (OGD), an semi-imputation method Attribute Efficient Ridge Regression (AERR) and an offline method Autoregressive Least Square impute (ARLS) as the baseline. The comparison is based on the results of extensive experiments on synthetic time series with various value settings of parameters and different categories of real data.

From the experimental results, we discover some important points: (1) AERR has the worst performance overally and it is especially weak on time series where the variation range is large, the change of observation value is dramatic and the missing rate is high; (2) despite of the better performance over AERR, OGD still has generally poor performance, and it is also sensitive to missing values in real data; (3) KF and YW have generally better performance. However, YW method is not suitable for time series with no seasonality or with unstable trend, and KF has poorer performance in the case of the noise is not Gaussian.

These conclusions implies that the imputation-absent methods are not ideal compared to the traditional imputation-based methods. However, the idea of reducing the adverse effect of bad predictions is still valuable.

Thus, future work includes novel sampling approaches for increasing the accuracy of imputation-absent prediction, effective parameter determination strategy to improve the precision as well as seek imputation-absent online time series prediction algorithms without sampling. In addition, although YW works generally well in many cases, its low efficiency limits its power on practical applications. Therefore, speeding up the computation process of YW is a possible direction to work on.

\bibliographystyle{unsrt}
\bibliography{references}

\end{document}